%% file: Formatting-Instructions-LaTeX-2026.tex
\documentclass[letterpaper]{article} 
\usepackage{aaai2026}  
\usepackage{times}  
\usepackage{helvet}  
\usepackage{courier}  
\usepackage[hyphens]{url}  
\usepackage{graphicx} 
\usepackage{natbib}  
\usepackage{caption} 
\usepackage{algorithm}
\usepackage{algorithmic}
\usepackage{bm,comment}
\usepackage{array}
\usepackage{tabularx}
\usepackage{multirow}
\usepackage{colortbl}
\usepackage{amsthm,amsmath,amssymb,amsfonts}
\usepackage{color}
\usepackage{latexsym}
\usepackage{arydshln}
\usepackage{graphicx}
\usepackage{subcaption}
\usepackage{enumitem}
\usepackage{cite}

\usepackage{newfloat}
\usepackage{listings}
\usepackage{cleveref}
\usepackage{appendix}
\usepackage{adjustbox}
\usepackage{booktabs}
\usepackage{newfloat}
\usepackage{listings}

\definecolor{weijia}{rgb}{0.21,0.49,0.74}

\newcommand{\e}{\text{e}}

\DeclareCaptionStyle{ruled}{labelfont=normalfont,labelsep=colon,strut=off} 
\floatstyle{ruled}
\newfloat{listing}{tb}{lst}{}
\floatname{listing}{Listing}
\title{BCE3S: Binary Cross-Entropy Based Tripartite Synergistic Learning \\ for Long-tailed Recognition}
\author {
    Weijia Fan   \textsuperscript{\rm 1, \rm 2, \rm 3, \rm 4}, 
    Qiufu Li     \textsuperscript{\rm 2, \rm 3}\thanks{Corresponding author.}, 
    Jiajun Wen   \textsuperscript{\rm 1, \rm 3}, 
    Xiaoyang Peng\textsuperscript{\rm 5}
}
\affiliations {
    \textsuperscript{\rm 1}College of Computer Science and Software Engineering, Shenzhen University \\
    \textsuperscript{\rm 2}School of Artificial Intelligence, Shenzhen University\\
    \textsuperscript{\rm 3}Guangdong Provincial Key Laboratory of Intelligent Information Processing, Shenzhen University \\
    
    \textsuperscript{\rm 4}CV:HCI Lab, Karlsruhe Institute of Technology \\
    \textsuperscript{\rm 5}School of Software Engineering, Sun Yat-sen University\\
    weijia.fan@partner.kit.edu, liqiufu@szu.edu.cn, wenjiajun@szu.edu.cn, pengxy77@mail2.sysu.edu.cn
}

\usepackage{bibentry}

\begin{document}

\maketitle

\begin{abstract}
For long-tailed recognition (LTR) tasks, high intra-class compactness and inter-class separability in both head and tail classes, as well as balanced separability among all the classifier vectors, are preferred.
The existing LTR methods based on cross-entropy (CE) loss not only struggle to learn features with desirable properties but also couple imbalanced classifier vectors in the denominator of its Softmax, amplifying the imbalance effects in LTR.
In this paper, for the LTR, we propose a binary cross-entropy (BCE)-based tripartite synergistic learning, termed BCE3S, which consists of three components:
(1) BCE-based joint learning optimizes both the classifier and sample features, which achieves better compactness and separability among features than the CE-based joint learning, by decoupling the metrics between feature and the imbalanced classifier vectors in multiple Sigmoid;
(2) BCE-based contrastive learning further improves the intra-class compactness of features;
(3) BCE-based uniform learning balances the separability among classifier vectors and interactively enhances the feature properties by combining with the joint learning.
The extensive experiments show that the LTR model trained by BCE3S not only achieves higher compactness and separability among sample features, but also balances the classifier's separability, achieving SOTA performance on various long-tailed datasets such as CIFAR10-LT, CIFAR100-LT, ImageNet-LT, and iNaturalist2018.
\end{abstract}
\begin{links}
    \link{Code}{https://github.com/wakinghours-github/BCE3S}
    
\end{links}
\section{Introduction}
\label{sec:intro}   
Though deep models trained on balanced datasets can surpass humans in visual recognition,
their performance is still unsatisfactory on the imbalanced datasets.
In the long-tailed datasets, the sample number decline significantly from the head to tail classes, which are more aligned with the data distribution in the real world, and the training of deep models on such datasets is easily dominated by the head classes, leading to imbalanced features and classifiers from the head to the tail classes, which consequently decreases the overall performance of the final models~\cite{Kang_2020_Decoupling,2022_maxnorm,xuan2024_aaai_dscl}.

    

\begin{table}[t]
    \centering
    \includegraphics[scale=0.48]{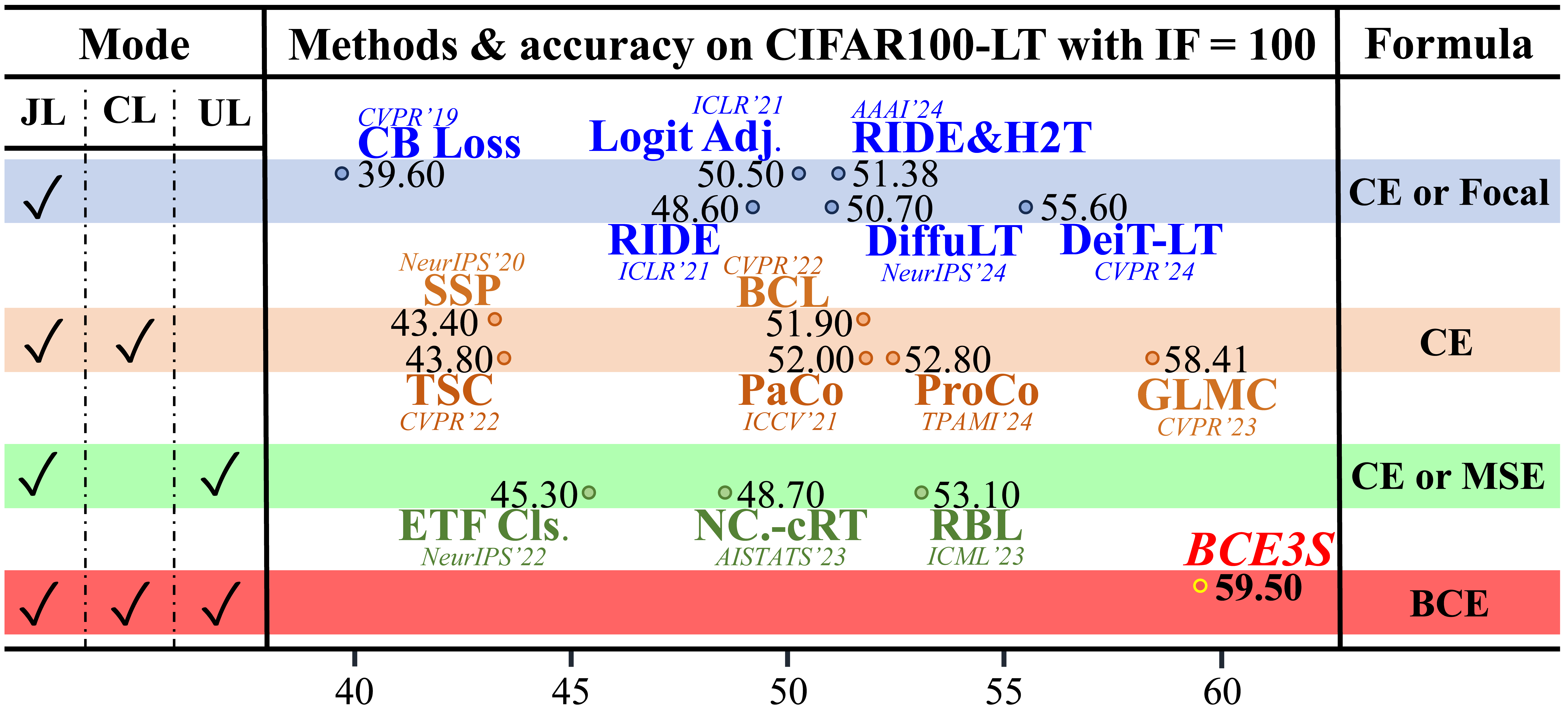}
    
    \caption{Comparison of BCE3S with previous LTR methods in terms of learning mode and performance. ``JL'', ``CL'', and ``UL'' are short for joint learning between sample features and classifier vectors, contrastive learning among features, and classifier's uniform separability learning. Our Only BCE3S integrates all the three learning modes and achieves the highest accuracy (i.e., $59.50\%$) on CIFAR100-LT with IF = 100. More results can be found in Table \ref{table:benchmark_on_cifar}. }
    \label{tab:comparison_of_bce3s}
    \vspace{-16pt}
\end{table}
In model training with cross-entropy (CE) loss, the classifier and sample features are jointly learned, while, on the long-tailed datasets, this learning mode is likely to fail to learn both well sample features and classifier.
Currently, the re-balancing techniques such as re-sampling~\cite{wallace2011class}, re-weighting~\cite{lin2017focal,cui2019class,2022_maxnorm}, and logit adjustment~\cite{Krishna2021logit_adjustment,zhao2022adaptive_logits_adjust,li2022logits_adjust_ce} have been developed and applied to improve the CE loss to balance the models' attention among different classes during the training.
Combined with the joint learning, contrastive learning~\cite{chen_simple_2020,he_momentum_2020} has also been used to enhance the feature properties and improve LTR~\cite{cui_2021_parametric_paco,kang2021exploring,du2023glmc,xuan2024_aaai_dscl}, by pulling the features of the same class closer and pushing those from different classes apart.

In LTR, in addition to enhancing sample features, it is also necessary to improve the discriminative capability of the classifiers learned on long-tailed datasets. Currently, there is no published work directly focusing on learning a well classifier for LTR.
However, according to neural collapse~\cite{papyan2020prevalence,zhu2021geometric}, the optimal classifier that CE can learn is an equiangular tight frame (ETF) classifier where any two classifier vectors exhibit uniform and maximum separability.
Then, customized ETF classifiers~\cite{yang_2022inducing_nc,kasarla_2022_maximum_NC,liu_2023_inducing} are designed before the training, while they may not align well with the final learned features.

{\color{black}
To improve LTR, a reasonable approach is to design a uniform separability learning framework for the classifier vectors and integrate it with joint learning and contrastive learning, creating a tripartite synergistic learning (TSL) paradigm to rebalance the features and classifiers during the model training.
However, in the LTR tasks, the commonly used CE loss couples imbalanced metrics from different classes on its Softmax's denominator, making it challenging to effectively suppress the imbalance effect by combining the various CE-based learning modes.}
{\color{black}As BCE (binary CE) decouples the metrics from different classes by adopting multiple Sigmoid, it can flexibly adjust these metrics and effectively suppress the imbalance effect. In fact, \citet{cui2019class} have experimentally demonstrated that BCE (i.e., Sigmoid cross-entropy in the reference) achieves better LTR than CE, yet the potential of BCE in LTR has not been fully explored.

Therefore, in this paper, we design a BCE-based tripartite synergistic learning approach for LTR, termed BCE3S, which integrates: (1) BCE-based joint learning between sample features and classifiers, (2) BCE-based contrastive learning among features, and (3) BCE-based uniform separability learning for the classifier vectors.
As Table~\ref{tab:comparison_of_bce3s} shows, with these three learning modes working in concert, BCE3S achieves the best LTR performance on CIFAR100-LT.

The main contributions of this paper are as follows:
\begin{enumerate}
    \item We introduce the concept of tripartite synergistic learning (TSL) and propose BCE3S, a BCE-based LTR method integrating the joint learning between features and classifier, contrastive learning among features, and uniform learning for classifier vectors.
    \item We explain in-depth that BCE-based uniform separability learning helps to train balanced classifier vectors, and analyze the advantage of BCE over CE in LTR.
    \item We conduct extensive experiments to evaluate the performance of BCE3S on LTR and find that compared to CE-based methods, BCE3S achieves better intra-class compactness and inter-class separability of sample features, and the BCE-based uniform learning can effectively balance the classifier's separability. BCE3S achieves SOTA LTR results on four long-tailed datasets.
\end{enumerate}

\section{Related Work}
\label{sec:related_work}

\subsection{Joint learning for both feature and classifier}
Various techniques have been proposed to improve the LTR performance of the popular CE-based joint learning.

The re-sampling technique~\cite{wallace2011class} adjusts the sample distributions via over-sampling the tail classes or under-sampling the head classes to mitigate the class imbalance of the long-tailed datasets, which complicates the model training.
The class-balanced loss~\cite{cui2019class} introduces a re-weighting factor inversely proportional to the sample numbers, which is adaptive to the CE, BCE, and focal losses based joint learning. In~\cite{Kang_2020_Decoupling}, the authors decoupled the learning of sample features and classifier, and they rectified the recognition decision boundaries on the head and tail classes by fine-tuning with different re-balancing strategies, including the classifier re-weighting.
DisAlign~\cite{zhang2021distribution} developed an adaptive calibration function to re-weight the learning of classifier vectors.
The methods of BaLMS~\cite{ren2020balanced_ms} and Logit Adj.~\cite{Krishna2021logit_adjustment} re-balance the model training by adjusting the logits according to the sample numbers.
Besides these techniques, distillation~\cite{rangwani_2024_deit}, diffusion~\cite{DiffuLT2024}, and collaborative learning~\cite{Towards2024Xu} have also enhanced the LTR performance.

The above works are mainly based on CE loss, while the published researches~\cite{cui2019class,wightman_2021_resnet,touvron_2022_deit} have shown that BCE also performs well in image recognition.
Especially, the result in~\cite{cui2019class} has experimentally shown that BCE has more potential than CE in LTR tasks, and LiVT \cite{xu_2023_VIT_BCE} is a BCE-trained Transformer for LTR.
However, these works do not in-depth explain the advantages of BCE over CE in LTR or further explore the application of BCE in this task.

\subsection{Contrastive learning on sample features}
Under the long-tailed scenario, PaCo~\cite{cui_2021_parametric_paco} incorporates a set of learnable class centers into contrastive learning, re-balancing the feature learning from a perspective of optimization. KCL~\cite{kang2021exploring} applies the same number of samples for each class in combining its positive pairs to learn balanced feature space by contrastive learning. SSD~\cite{li_2021_self_SSD} employs a self-distillation framework to enhance the information exploitation in the tail class.
BCL~\cite{zhu_2022_balanced_BCL} introduces class-averaging and class-complement to balance the gradient contribution in the contrastive learning. 
GLMC~\cite{du2023glmc} enhances the robustness of LTR model via mitigating the head class bias by designing a re-weighting loss.
\citet{xuan2024_aaai_dscl} propose decoupled supervised contrastive loss and patch-based self-distillation to alleviate the imbalance effect in LTR.
ProCo~\cite{du_2024_probabilistic_proco} models the feature space using von Mises-Fisher distribution to eliminate the limitation of contrastive learning in requiring a large amount of samples in LTR tasks.
These contrastive learning methods are designed on Softmax, measuring the relative value of feature similarity and amplifying the imbalance effect in LTR.

\subsection{Uniform separability learning for classifier}
The works of neural collapse~\cite{papyan2020prevalence,zhu2021geometric} show that when training a recognition model on a balanced dataset and the CE loss reaches its minimum, the classifier vectors form an equiangular tight frame (ETF), where uniform and maximum separability exists between any two classifier vectors.
However, on an imbalanced dataset, if the imbalance factor (IF) is too large, the classifier vectors of the tail classes will converge and collapse to the same vector~\cite{fang_2021_exploring_nc}, losing their separability.

To keep the classifier's uniform separability, \citet{yang_2022inducing_nc}, \citet{liu_2023_inducing}, and \citet{kasarla_2022_maximum_NC} customize ETF classifiers before the LTR model training, which do not participate in the training and maintain the uniform separability; RBL \cite{Peifeng_2023_nc_LTR} adopts a learnable orthogonal matrix to rotate the ETF classifier, adjusting its direction based on the sample features.
However, these customized ETF classifiers struggle to align with the learned features.
In this paper, we propose BCE-based uniform learning, helping to learn balanced classifiers aligned with features.

\section{Method}
\label{sec:method}

\subsection{Preliminary}\label{sec:test}
Let $\mathcal{D}=\bigcup_{k=1}^K \mathcal{D}_k$ be a dataset from $K$ classes,
where $\mathcal{D}_k$ contains the samples from the $k$-th class,
and $n_k=|\mathcal{D}_k|$ denotes its sample number.
Suppose $\mathcal D$ is an imbalanced and long-tailed dataset,
and its $K$ subsets have been sorted by the sample numbers in descending order, i.e., $n_k\geq n_\ell$
for $\forall k < \ell$, then $n_1 / n_K$ denotes the imbalance factor (IF).

In recognition task, for any sample $\bm X\in\mathcal D$,
an deep model $\mathcal{M}(\cdot)$ first extracts its feature $\bm x = \mathcal M(\bm X) \in \mathbb R^d$,
then, a linear classifier $\mathcal C = \{(\bm w_j, b_j)\}_{j=1}^K$ converts it into $K$ metrics $\{\bm w_j^T \bm x + b_j\}_{j=1}^K$,
which are applied to predict the sample's label, $\hat k = \arg\max_j\{\bm w_j^T \bm x + b_j\}_{j=1}^K$.

In training, for a sample from the $k$-th subset $\mathcal D_k, \forall k$,
we denote it as $\bm X^{(k)}$ and its feature as $\bm x^{(k)} = \mathcal M(\bm X^{(k)})$.
In the previous works
\cite{ren2020balanced_ms,li2022logits_adjust_ce,2022_maxnorm,du2023glmc,xuan2024_aaai_dscl,wang2021longtailed_RIDE,chen_2023_area},
to learn the better model $\mathcal M$ and classifier $\mathcal C = \{(\bm w_j, b_j)\}_{j=1}^K$,
Softmax was first applied to compute the probabilities
$\big\{ \frac{\exp(\bm{w}_j^T \bm{x}^{(k)} + b_j)}{\sum_{\ell=1}^K \exp(\bm{w}_\ell^T \bm{x}^{(k)} + b_\ell)} \big\}_{j=1}^K$
that $\bm X^{(k)}$ belongs to each class,
then, cross-entropy was used to compute the CE loss,
\begin{align}
\label{eq:sample_to_class_CE}
    L_{\text{ce}}^{\text{(sc)}}(\bm x^{(k)}) = -\log\Big(\frac{\exp{(\bm w_k^T \bm x^{(k)} + b_k)}}{\sum_{\ell=1}^{K}\exp{(\bm w_\ell^T \bm x^{(k)} + b_\ell)}}\Big).
\end{align}
When applying CE loss to train models, the classifier vectors $\{\bm w_k\}$ and the sample features $\{\bm x^{(k)}\}$ are jointly trained and learned.
The re-balancing techniques, such as re-sampling and re-weighting, have been used to improve CE loss and enhance its performance on long-tailed recognition (LTR).

Contrastive learning based on Softmax has also been introduced into LTR, comparing features in a projection space,
\begin{align}
\label{eq:sample_to_sample_CE}
L_{\text{ce}}^{\text{(ss)}}(\bm x^{(k)}) = -\log\Big(\frac{\exp\big(\frac{1}{\tau}\cos(\bm z^{(k)},\bm z_*^{(k)})\big)}
            {\sum_{\ell=1}^{N}\exp\big(\frac{1}{\tau}\cos(\bm z^{(k)},\bm z_*^{(\ell)})\big)}\Big),
\end{align}
where $\bm z^{(k)} = \mathcal P(\bm x^{(k)}), \bm z^{(k)}_* = \mathcal P(\bm x^{(k)}_*)$, and $\mathcal P$ is a non-linear projection operator \cite{chen_simple_2020}.
In Eq. (\ref{eq:sample_to_sample_CE}), $\tau$ is a temperature factor, $\cos(\bm z,\bm z_*) = \frac{\langle\bm z, \bm z_*\rangle}{\|\bm z\|\|\bm z_*\|}$ denotes the cosine similarity of any two vectors $\bm z, \bm z_*\in\mathbb R^{d'}$, and $\{\bm z_*^{(k)}\}_{k=1}^K$ are temporary features from the $K$ classes, which could be saved in memory bank during the training.

\textcolor{black}{
According to neural collapse~\cite{papyan2020prevalence,zhu2021geometric}, the optimal classifier that CE can learn is the one whose classifier vectors $\{\bm w_k\}_{k=1}^K$ form an ETF which has uniform and maximum separability, i.e., satisfying $\cos(\bm w_k, \bm w_{k'}) = - \frac{1}{K-1}$ and $\|\bm w_k\| = \|\bm w_{k'}\|$ for $\forall k\neq k'$.
}
The customized ETF classifiers~\cite{yang_2022inducing_nc,liu_2023_inducing} have been used to design LTR methods.

\begin{figure}[t]
    \centering
    \includegraphics[width=1.0\linewidth]{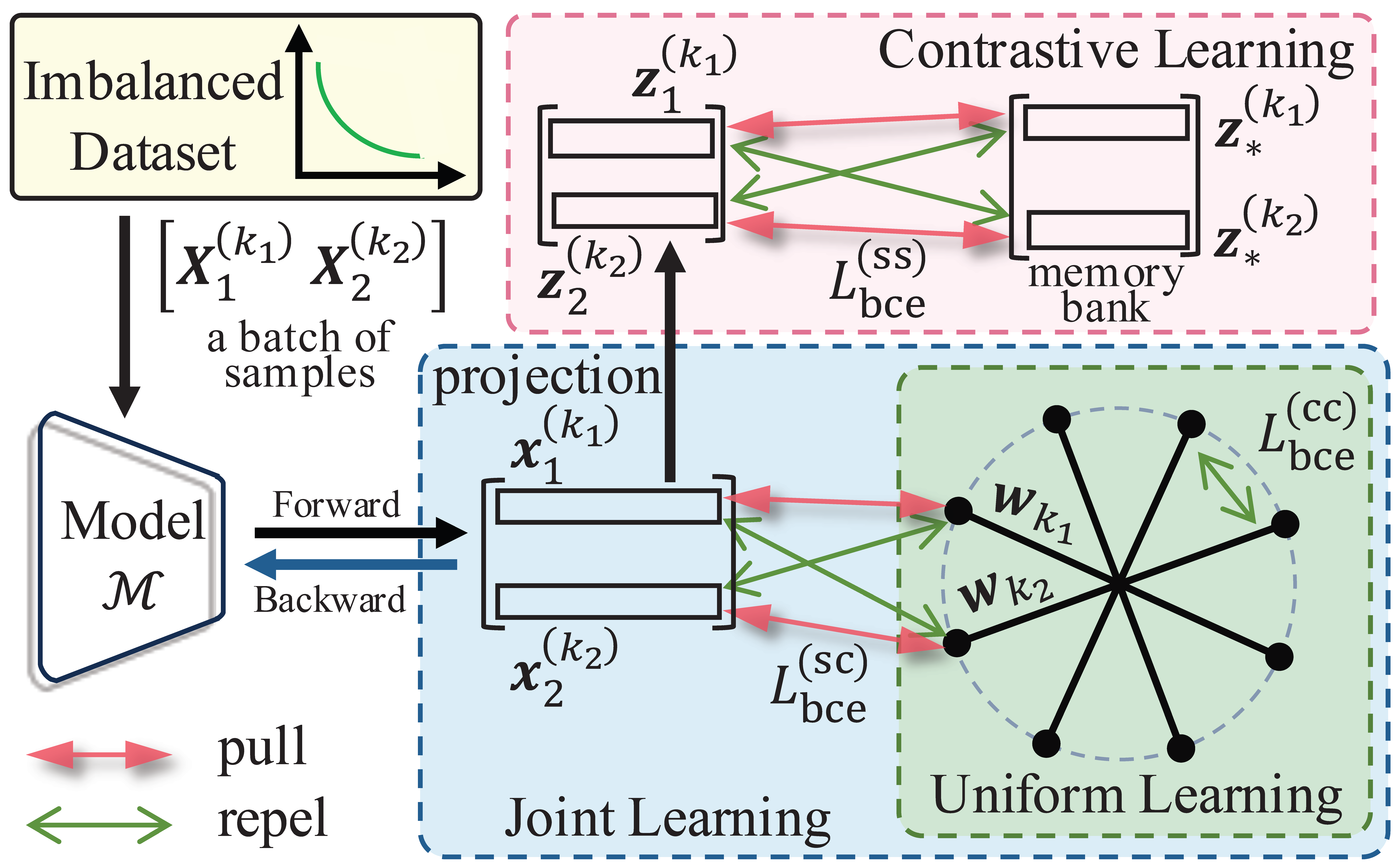}

    \caption{Pipeline of BCE3S, integrating all the three learning modes, i.e., joint learning (Eq.~(\ref{eq:sample_to_class_BCE})) between sample feature and classifier, contrastive learning (Eq.~(\ref{eq:sample_to_sample_BCE})) among features, and classifier's uniform separability learning (Eq.~(\ref{eq:class_to_class_learning})).}
    \label{fig:pipline_tri_learning}
    \vspace{-8pt}
\end{figure}


\subsection{BCE-based tripartite synergistic learning}\label{sec:bce3s}
We propose the BCE-based tripartite synergistic learning (TSL), i.e., BCE3S, to integrate the above three learning modes. BCE3S consists of 
BCE-based joint learning $\textcolor[rgb]{0,0,0}{L^{(\text{sc})}_{\text{bce}}}$, 
BCE-based contrastive learning $\textcolor[rgb]{0,0,0}{L^{(\text{ss})}_{\text{bce}}}$, 
and BCE-based uniform learning $\textcolor[rgb]{0,0,0}{L^{(\text{cc})}_{\text{bce}}}$. 
Fig. \ref{fig:pipline_tri_learning} shows its pipeline.

{\textbf{Joint learning.}}\quad For a sample feature $\bm x^{(k)}$ from the $k$-th class, the joint learning $L^{(\text{sc})}_{\text{bce}}$ measures its similarities with the normalized classifier vectors using BCE formula,
\begin{align}
        \label{eq:sample_to_class_BCE}
        L_{\text{bce}}^{(\text{sc})}(\bm x^{(k)}) &= \log\big(1 + \exp(- {\bm w}_k^T \bm x^{(k)} - b_k)\big) \nonumber\\
        &~ + \sum_{j=1, j\neq k \atop p_j<r}^{K}\log\big(1 + \exp({\bm w}_j^T \bm x^{(k)} + b_j)\big),
\end{align}
where $\|\bm w_j\| = 1,\,\forall j$.
In Eq. (\ref{eq:sample_to_class_BCE}), the normalization of the classifier vectors could prevent their gradients from being dominated by the head class in training. which helps to learn more uniform classifier vectors in norm~\cite{Kang_2020_Decoupling}.

In $L^{(\text{sc})}_{\text{bce}}$, $r\in(0,1]$ is a re-sampling parameter for the negative sample-to-class metrics $\{{\bm w}_j^T \bm x^{(k)}\}_{j\neq k}$,
and $p_j$ is randomly sampled from the uniform distribution $U(0,1)$.

{\textbf{Contrastive learning.}}\quad
For any sample feature $\bm x^{(k)}$ from class $k$, similar to \cite{chen_simple_2020}, $L^{(\text{ss})}_{\text{bce}}$ implements the contrastive learning in a projection space.
A non-linear projection operator $\mathcal P$ is first applied to transform the sample feature into $\bm z^{(k)} = \mathcal P(\bm x^{(k)})\in \mathbb R^{d'}$,
and
\begin{align}
\label{eq:sample_to_sample_BCE}
L_{\text{bce}}^{\text{(ss)}}(\bm x^{(k)}) &= \log\Big(1+\exp\big(-\frac{1}{\tau}\cos\big(\bm z^{(k)}, \bm z_*^{(k)}\big)\big)\Big) \nonumber\\
+ & \sum_{j=1 \atop j\neq k}^{K}\log\Big(1+\exp\big(\frac{1}{\tau}\cos\big(\bm z^{(k)}, \bm z_*^{(j)}\big)\big)\Big),
\end{align}
where $\big\{\bm z_*^{(j)} = \mathcal P(\bm x_*^{(j)})\big\}_{j=1}^K$ are projections of $K$ sample features $\big\{\bm x_*^{(j)}\big\}_{j=1}^K$ stored in a memory bank.

{\textbf{Uniform separability learning for classifier.}}\quad
In the LTR tasks, the joint learning $L_{\text{bce}}^{(\text{sc})}$ tends to learn more discriminative classifier vectors for the head classes but less discriminative ones for the tail classes,
and the contrastive learning $L_{\text{bce}}^{(\text{ss})}$ focuses to the sample features, which would aggravate the imbalanced effect.
To balance the discriminative property of the classifier vectors for the different classes, we design a uniform separability learning for the classifier,
\begin{align}
    \label{eq:class_to_class_learning}
    L_{\text{bce}}^{(\text{cc})}(\bm w_k) &= \log \Big(1 + \exp\big(- {\bm w}^T_k  {\bm w}_k\big)\Big) \nonumber\\
    & \quad + \sum_{j=1 \atop j\neq k}^{K}\log\Big(1 + \exp\big( {\bm w}_k^T  {\bm w}_j\big)\Big),
\end{align}
where the positive term equals to $\log(1+\e^{-1})$ as $\|\bm w_k\|^2 = 1$,
which is a constant and was omitted in our experiments.
The uniform learning $L_{\text{bce}}^{(\text{cc})}$ tries to maximize the separability among the classifier vectors $\{\bm w_k\}_{k=1}^K$,
which helps to learn classifier with uniform and maximum separability (i.e., ETF classifier) and aligning with the final sample features by working together with the joint learning in BCE3S.

{\textbf{Training pipeline.}}\quad
In the experiments, we comprehensively learn the sample features and classifier vectors using BCE3S.
As Fig.~\ref{fig:pipline_tri_learning} shows, in each iteration during the training, a batch of samples $[\bm X_i^{(k_i)}]_{i=1}^B$ are fed into the model $\mathcal M$ to extract their features $[\bm x_i^{(k_i)}]_{i=1}^B$,
where $B$ is batch size.
A tripartite synergistic loss is computed to train the model,
\begin{align}
    \label{eq:tri_learning_loss}
    L^{(\text{tri})}_{\text{bce}}\big([\bm x_i^{(k_i)}]\big) &= \frac{1}{B}\sum_{i=1}^B L^{(\text{sc})}_{\text{bce}}(\bm x_i^{(k_i)})
                                   + \frac{\lambda_{\text{ss}}}{B}\sum_{i=1}^B L^{(\text{ss})}_{\text{bce}}(\bm x_i^{(k_i)}) \nonumber\\
                                  & + \frac{\lambda_{\text{cc}}}{K}\sum_{k=1}^K L^{(\text{cc})}_{\text{bce}}(\bm w_{k}),
\end{align}
where $\lambda_{\text{ss}}$ and $\lambda_{\text{cc}}$ are weight factors.

Although we design BCE3S, it should be noted that other losses, such as CE and MSE, can also be integrated within the TSL framework. However, for the LTR tasks, BCE not only inherits the advantage of fast convergence from CE but also suppresses the imbalance effects.

\subsection{Analysis for tripartite synergistic learning (TSL)}\label{sec:analysis_BCE3S}
We analyze BCE3S in balancing the separability among the different classes and enhancing the feature's properties.

{\textbf{Classifier during the training.}}\quad
Batch algorithm and back propagation are commonly used in the model training.
In a iteration, with a batch of features $[\bm x_i^{(k_i)}]_{i=1}^B$,
the normalized classifier vector $ {\bm w}_k$ is updated via its gradient, i.e.,
\begin{align}
{\bm w}_k \leftarrow{\bm w}_k - \eta \frac{\partial L_{\text{bce}}^{(\text{tri})}\big([\bm x_i^{(k_i)}]\big)}{\partial {\bm w}_k},~~\forall\,k,
\end{align}
where $\eta$ is the learning rate,
and
\begin{align}
 \frac{\partial L_{\text{bce}}^{(\text{tri})}\big([\bm x_i^{(k_i)}]\big)}{\partial {\bm w}_k} =
 -\frac{1}{B}\bigg(\sum_{i=1\atop k_i=k}^B\underbrace{\sigma(-{\bm w}_k^T \bm x_i^{(k_i)}) \bm x^{(k_i)}_i}_{\text{pulling term}} &\nonumber \\
  - \sum_{i=1\atop k_i\neq k}^B \underbrace{\sigma({\bm w}_k^T \bm x_i^{(k_i)})\bm x_i^{(k_i)}}_{\text{repelling term}} \bigg)
 + \frac{1}{K}\sum^{K}_{j=1 \atop j\neq k }\underbrace{\sigma({\bm w}_k^T{\bm w}_j){\bm w}_j}_{\text{interactive  term}},& \label{eq_gradient_w_k}
\end{align}
where $\sigma$ is Sigmoid.
In Eq. (\ref{eq_gradient_w_k}), the pulling terms and repelling terms are derived from the joint learning $L_{\text{bce}}^{\text{(sc)}}$,
and they pull the classifier vector $\bm w_k$ using the sample features $\bm x_i^{(k_i=k)}$ from the same class
and repel it using the ones $\bm x_i^{(k_i\neq k)}$ from the different classes.
Due to the higher probability that the batch contains samples from the head classes,
the classifier vectors of the head classes are prone to be repelled by the samples of other head classes in different directions,
while the tail classifier vectors are not easily to be repelled by the samples from the other tail classes,
which results in high separability among the head classifier vectors and the low ones among the tail classifier vectors.
The tail classifier vectors are also easily repelled by the head class samples, which would lead to higher separability between classifier vectors of the head and tail classes,
because the repulsion from the head class is more concentrated on the tail classifier vectors.
Totally, the joint learning $L_{\text{bce}}^{\text{(sc)}}$ will learn a imbalanced classifier on LTR, which will affect the feature learning and result in imbalanced features across classes.

In contrast, the uniform learning $L^{(\text{cc})}_{\text{bce}}$ leads to the interactive term in Eq.~(\ref{eq_gradient_w_k}).
For each $\bm w_k, \forall k$, in every batch, the interactive term provides $K-1$ repelling forces from other classifier vectors,
which directly, uniformly, and consistently maximize the separability among the $K$ classifier vectors.
Thereby, $L^{(\text{cc})}_{\text{bce}}$ helps to learn a balanced classifier on LTR.
Combined with $L^{(\text{sc})}_{\text{bce}}$, the balanced classifier will align with the feature learning, and re-balance the features.

The contrastive learning $L^{(\text{ss})}_{\text{bce}}$ enhances the feature learning in the training, which does not directly update the classifier.
However, when combined with joint learning $L^{(\text{sc})}_{\text{bce}}$, it can further exacerbate the imbalance of classifier separability through imbalanced sample features.

{\textbf{CE vs. BCE in LTR.}}\quad
On the LTR, both CE-based and BCE-based joint learning, $L_{\text{ce}}^{(\text{sc})}$ and $L_{\text{bce}}^{(\text{sc})}$, result in imbalanced classifier vectors among the head and tail classes,
while they perform diversely on the sample features.

For a sample feature $\bm x^{(k)}$, it is updated by
\begin{equation}
    \bm x^{(k)} \leftarrow \bm x^{(k)} - \eta \frac{\partial L^{\text{(sc)}}_{\mu}\big(\bm x^{(k)}\big)}{\partial \bm x^{(k)}},
\end{equation}
where $\mu \in \{\text{ce}, \text{bce}\}$ denotes CE- or BCE-based joint learning, and $\eta$ is the learning rate.
The gradients $\frac{\partial L_{\mu}^{(\text{sc})}(\bm x^{(k)})}{\partial \bm x^{(k)}}$ have the similar form in the CE or BCE,
\begin{align}
      \underbrace{-\big(1-\mathtt{Act}_{\mu}({\bm w}_k ^T\bm x^{(k)})\big){\bm w}_k}_{\text{pulling}} + \sum_{j=1 \atop j\neq k}^K\underbrace{\mathtt{Act}_{\mu}({\bm w}_j ^T\bm x^{(k)}) {\bm w}_j}_{\text{repelling}}
\end{align}
where, for $\forall j$,
\begin{align}
 &\mathtt{Act}_{\text{ce}}({\bm w}_j ^T\bm x^{(k)})  = \frac{\exp({\bm w}_j^T \bm x^{(k)} + b_j)}{\sum_{\ell=1}^K \exp({\bm w}_\ell^T \bm x^{(k)} + b_\ell)},\\
 &\mathtt{Act}_{\text{bce}}({\bm w}_j ^T\bm x^{(k)}) = \sigma({\bm w}_j ^T\bm x^{(k)}). 
\end{align}
In the feature updating, $L_{\text{ce}}^{\text{(sc)}}$ and $L_{\text{bce}}^{\text{(sc)}}$ utilize the exponential inner products of the feature $\bm x^{(k)}$ with $K$ classifier vectors $\{{\bm w}_j \}_{j=1}^K$ to compute one pulling term and $K-1$ repelling terms, while the imbalanced classifier vectors $\{{\bm w}_j \}_{j=1}^K$ will slow down the feature learning.

For CE-based joint learning, $L_{\text{ce}}^{\text{(sc)}}$, in computing each pulling or repelling term, the $K$ imbalanced exponential inner products are coupling on the denominator of Softmax $\mathtt{Act}_{\text{ce}}$, thereby, the imbalanced effect is \textbf{injected again} into the feature learning.
In contrast, BCE decouples the metrics between the feature with $K$ classifier vectors, and only one classifier vector is applied in computing one pulling or repelling term by Sigmoid $\mathtt{Act}_{\text{bce}}$.
Therefore, BCE will achieve more balanced features across classes.

Similarly, BCE-based contractive learning and uniform separability learning, $L_{\text{bce}}^{\text{(ss)}}$ and $L_{\text{bce}}^{\text{(cc)}}$, could respectively result in better classifier and features than the CE-based ones, $L_{\text{ce}}^{\text{(ss)}}$ and $L_{\text{ce}}^{\text{(cc)}}$ (see supplementary for their formulas).

During the training, compared to CE, BCE adds at most $K-1$ logarithmic operations, which is negligible.
In the testing or inference, it does not incur any additional cost.

\section{Experiments and Results}
\label{sec:experiment}

\textcolor{black}{We evaluate BCE3S on four long-tailed datasets: CIFAR10-LT/CIFAR100-LT~\cite{yue2016deep}, ImageNet-LT~\cite{liu2022open_imagenet_lt}, and iNaturalist2018~\cite{van2018inaturalist}.
The CIFAR variants have IF of 100, 50, and 10, while ImageNet-LT and iNaturalist2018 have IF of 256 and 500, respectively. 
We train ResNets on their imbalance training set and evaluate them on the balance test set. Following~\cite{Kang_2020_Decoupling,2022_maxnorm,du_2024_probabilistic_proco}, we report total accuracy on the entire test set and three subsets: \texttt{Many} ($>100$ samples per class), \texttt{Medium} ($\leq 100$ and $\geq 20$ samples per class), and \texttt{Few} ($<20$ samples per class). More details about datasets and training can be found in supplementary.}

\begin{table}
    \centering
    \small
    \setlength{\tabcolsep}{0.88mm}{
        \begin{tabular}{cccccc|cccc}
        \hline
        \multicolumn{6}{c|}{Methods} & \multirow{2}{*}{\texttt{Many}} & \multirow{2}{*}{\texttt{Med.}} & \multirow{2}{*}{\texttt{Few}} & \multicolumn{1}{c}{\multirow{2}{*}{All}}  \\
        \cline{1-6}
        $L_{\text{ce}}^{\text{(sc)}}$ & $L_{\text{ce}}^{(\text{ss})}$& $L_{\text{ce}}^{(\text{cc})}$& $L_{\text{bce}}^{\text{(sc)}}$& $L_{\text{bce}}^{(\text{ss})}$ & $L_{\text{bce}}^{(\text{cc})}$ & &   &      & \multicolumn{1}{c}{}    \\
        \hline
        \checkmark  & ~            & ~           & ~           & ~              & ~            & 82.29  & 51.37  & 15.67  & 51.48  \\
        ~           & ~            & ~           & \checkmark  & ~              & ~            & 81.11  & 55.06  & 17.40  & 52.88  \\\hdashline
        \checkmark  & \checkmark   & ~           & ~           & ~              & ~            & 84.09  & 54.80  & 17.53  & 53.87  \\
        \checkmark  & ~            & ~           & ~           & \checkmark     & ~            & \textbf{84.17}  & 55.20  & 17.90  & 54.15  \\
        ~           & \checkmark   & ~           & \checkmark  & ~              & ~            & 83.37  & 55.83  & 19.87  & 54.68  \\
        ~           & ~            & ~           & \checkmark  & \checkmark     & ~            & 82.74  & 56.57  & 20.63  & 54.95  \\\hdashline
        \checkmark  & ~            & \checkmark  & ~           & ~              & ~            & 82.31  & 51.83  & 17.20  & 52.11  \\
        \checkmark  & ~            & ~           & ~           & ~              & \checkmark   & 81.23  & 53.69  & 18.17  & 52.67  \\
        ~           & ~            & \checkmark  & \checkmark  & ~              & ~            & 81.57  & 55.51  & 17.93  & 53.36  \\
        ~           & ~            & ~           & \checkmark  & ~              & \checkmark   & 81.03  & 56.51  & 19.20  & 53.90  \\\hdashline
        \checkmark  & \checkmark   & \checkmark  & ~           & ~              & ~            & 83.97  & 54.54  & 18.87  & 54.14  \\
        \checkmark  & ~            & ~           & ~           & \checkmark     & \checkmark   & 83.77  & 54.49  & 18.67  & 53.99  \\
        ~           & \checkmark   & \checkmark  & \checkmark  & ~              & ~            & 83.77  & 56.17  & 21.40  & 55.40  \\
        ~           & ~            & ~           & \checkmark  & \checkmark     & \checkmark   & 83.34  & \textbf{57.09}  & \textbf{22.80}  & \textbf{55.99}  \\
        \hline
        \end{tabular}
    }
    \caption{Ablation study for the proposed BCE3S and the CE based counterparts on CIFAR100-LT (IF $=100$). The formulas of CE-based contrastive learning and uniform learning, $L_{\text{ce}}^{(\text{ss})}$ and $L_{\text{ce}}^{(\text{cc})}$, are presented in supplementary. }
    \label{table:ablation_study}
    \vspace{-16pt}
\end{table}

\subsection{Ablation Study}
{\textbf{LTR results.}}\quad
We conduct ablation study for the proposed BCE3S on CIFAR100-LT with IF $=100$ using ResNet32, and Table \ref{table:ablation_study} presents the results.
We first take the CE-based joint learning $L_{\text{ce}}^{(\text{sc})}$ as baseline.
When training the model using BCE-based joint learning $L_{\text{bce}}^{(\text{sc})}$,
though it reduces the accuracy on the \texttt{Many} subset, it improves the accuracy on the \texttt{Medium} and \texttt{Few} subsets,
and increasing the overall accuracy from $51.48\%$ to $52.88\%$,
indicating that BCE pay more attention to the tail classes and improves the total accuracy at the cost of reduced accuracy on the head classes.


\textcolor{black}{
Based on $L_{\text{ce}}^{(\text{sc})}$, we compare CE- and BCE-based contrastive learning, $L_{\text{ce}}^{(\text{ss})}$ and $L_{\text{bce}}^{(\text{ss})}$. As Table \ref{table:ablation_study} shows, $L_{\text{bce}}^{(\text{ss})}$ improves accuracy across all subsets, increasing total accuracy from $53.87\%$ to $54.15\%$, and further to $54.95\%$ when combined with $L_{\text{bce}}^{(\text{sc})}$. For uniform learning, $L_{\text{bce}}^{(\text{cc})}$ enhances the performance on \texttt{Medium} and \texttt{Few} despite reducing that on \texttt{Many}, yielding better overall performance ($52.67\%$), which increases to $53.90\%$ when combined with $L_{\text{bce}}^{(\text{sc})}$;
meanwhile, it always perform better than CE-based one $L_{\text{ce}}^{(\text{cc})}$.
}

We finally compare different tripartite synergistic learning approaches based on CE and BCE, and achieve the best total LTR accuracy ($55.99\%$) using the proposed BCE3S.

{\textbf{Separability and compactness.}}\quad
To analyze the models trained with various learning methods, for any class $k$,
we define intra-class compactness $\mathcal E^{\text{(x,com)}}_k$, inter-class separability $\mathcal E^{\text{(x,sep)}}_k$ among features,
and separability $\mathcal E^{\text{(w,sep)}}_k$ among the classifier vectors;
as their names imply, $\mathcal E^{\text{(x,com)}}_k$ measures the compactness of features within the $k$-th class, $\mathcal E^{\text{(x,sep)}}_k$ measures the separability between the features from $k$-th class and that from other classes, and $\mathcal E^{\text{(w,sep)}}_k$ measures the separability between the $k$-th classifier vector with other ones, seeing supplementary for their definitions. The higher values suggest the better separability or compactness of the classifier vectors or features.
Fig.~\ref{fig:CIFAR100_fu_comparsion_matrices} shows the results of these metrics for CE- and BCE-based methods, as well as their mean and standard deviation across different classes.

As the first row in the figure shows, the four CE-based methods achieve similar intra-class compactness, with their average values all approximately $82$, and a noticeable imbalance is shown from the head to tail classes. In contrast, for the BCE-bsed methods, the average compactness of $L_{\text{bce}}^{\text{(sc)}}$ reaches to $86.02$.
With the addition of $L_{\text{bce}}^{\text{(ss)}}$ and $L_{\text{bce}}^{\text{(cc)}}$, the compactness gradually increase to $95.47$, while the imbalance across different classes gradually decreases and standard deviation of the compactness decreases from $5.55$ to $1.81$.
As the second row shows, compared to the CE-based methods, the BCE-based ones achieve higher inter-class separability for sample features,
with the highest mean of separability increases from $49.71$ ($L_{\text{ce}}^{\text{(sc)}} + L_{\text{ce}}^{\text{(ss)}}$) to $50.84$ ($L_{\text{bce}}^{\text{(sc)}} + L_{\text{bce}}^{\text{(ss)}} + L_{\text{bce}}^{\text{(cc)}}$).
Totally, compared to CE-based methods, BCE3S enhances the feature properties and balances the property difference on the head and tail classes.

As the third row of Fig. \ref{fig:CIFAR100_fu_comparsion_matrices} shows, the average separability of classifier vectors for the various methods is similar, close to $50.50$, while they have significant differences across the different classes.
The classifier separability of the two joint learning $L_{\text{ce}}^{\text{(sc)}}$ and $L_{\text{bce}}^{\text{(sc)}}$ has significant imbalance from the head to tail classes.
When the CE- and BCE-based contrastive learning are added, the imbalance is respectively amplified, while the two kinds of uniform learning can reduce the imbalance effect.
In particular, the BCE one significantly suppresses the separability imbalance, resulting in a standard deviation of only $0.106$ for $L_{\text{bce}}^{\text{(sc)}} + L_{\text{bce}}^{\text{(cc)}}$.

These results indicate that BCE3S can not only enhance the feature properties, but also effectively improve the balance of features and classifier for the LTR models.

\begin{figure}[t]
    \centering
    \includegraphics*[scale=0.235, viewport=8 30 538 398]{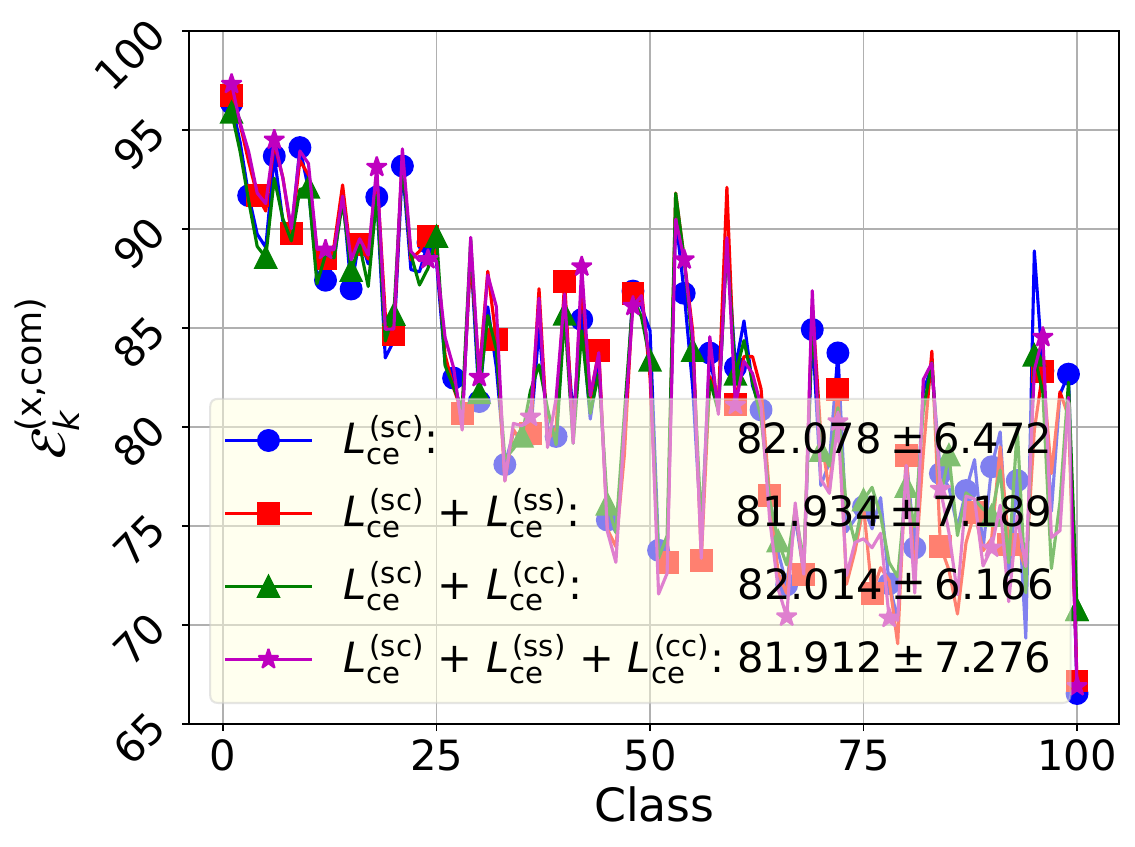}\hspace{0pt}
    \includegraphics*[scale=0.235, viewport=81 30 538 398]{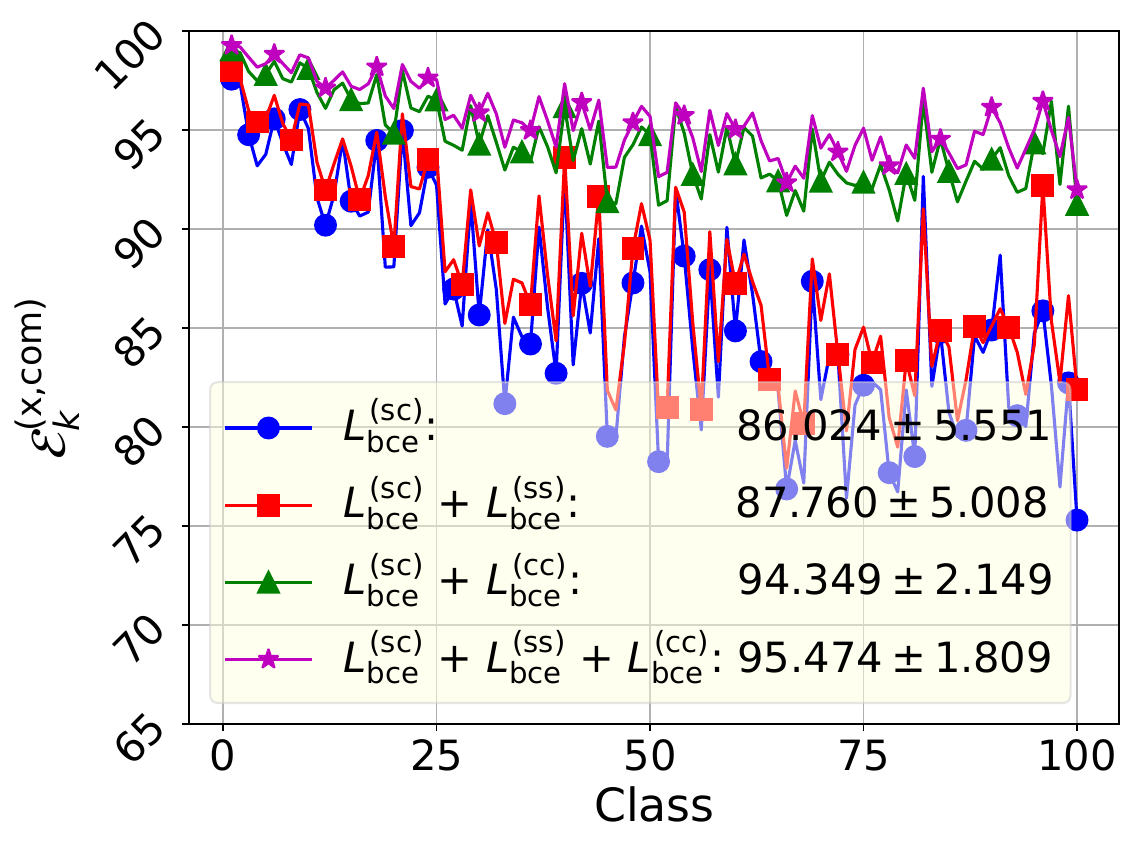}\\
    \includegraphics*[scale=0.235, viewport=8 30 545 398]{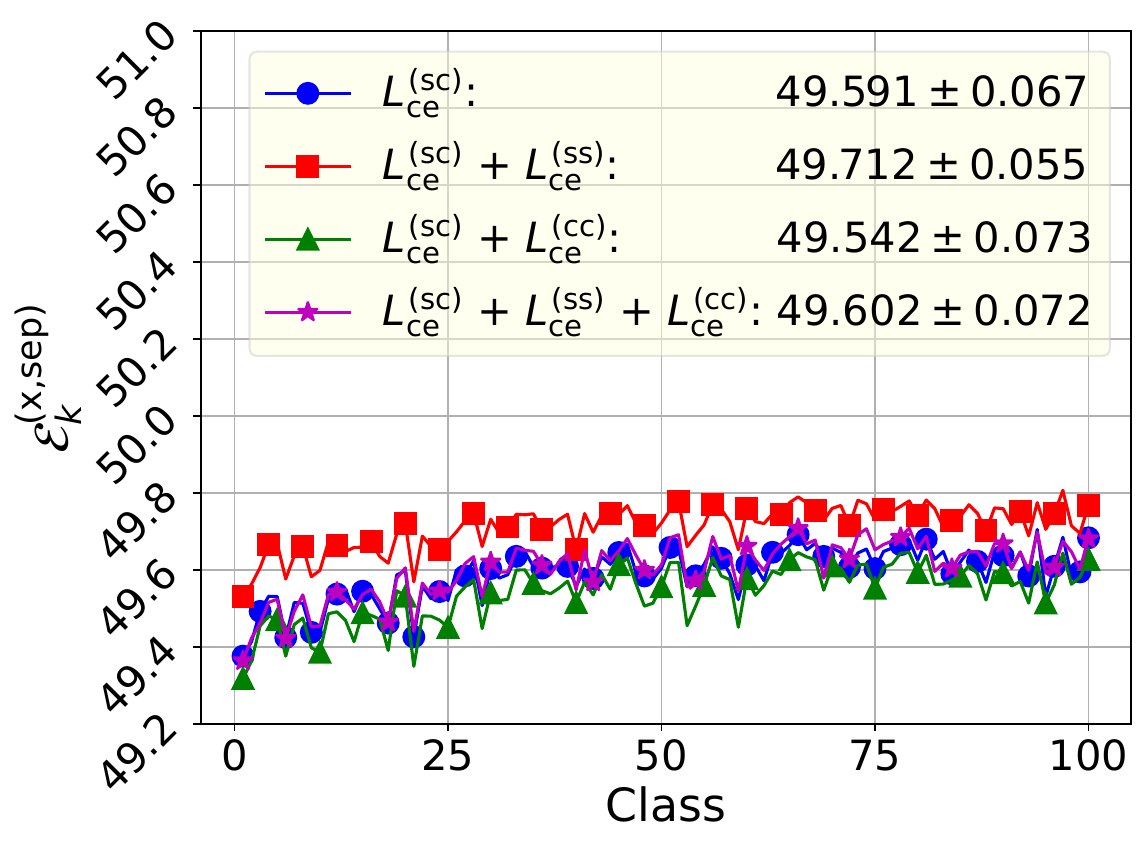}\hspace{0pt}
    \includegraphics*[scale=0.235, viewport=85 30 545 398]{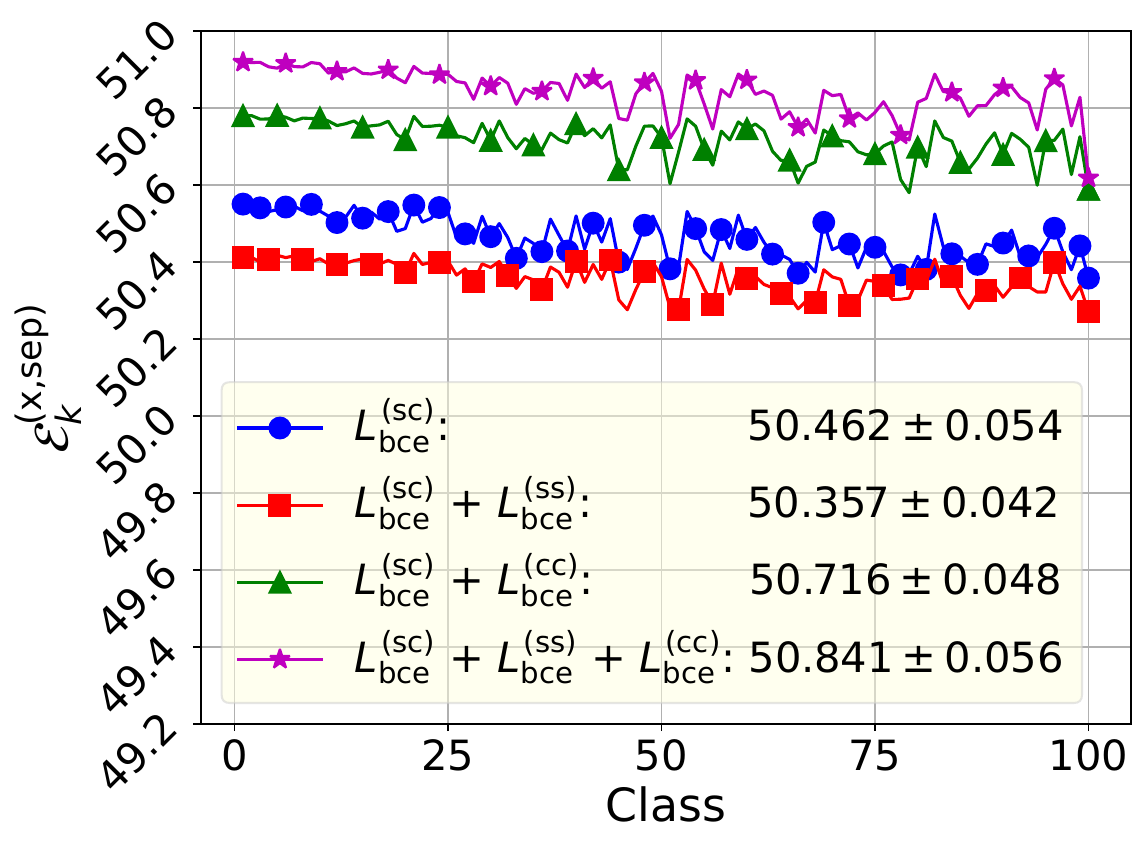}\\
    \includegraphics*[scale=0.235, viewport=8 2 545 398]{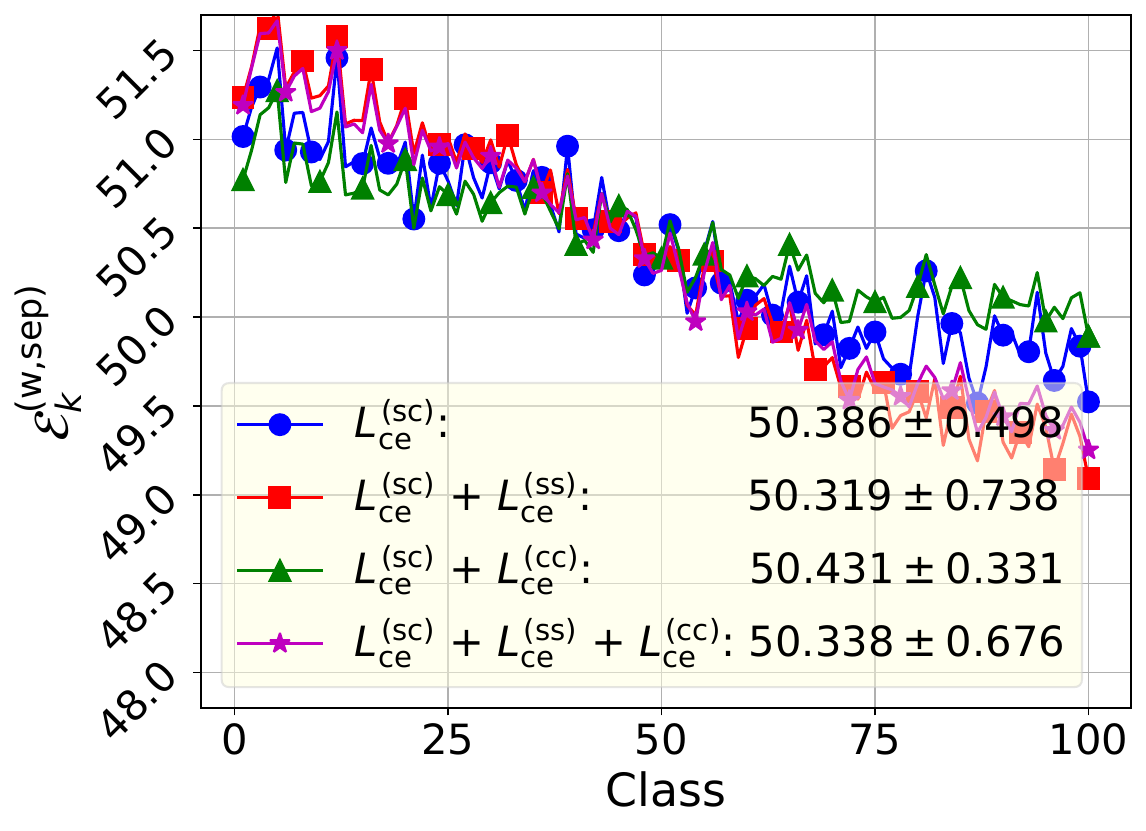}\hspace{0pt}
    \includegraphics*[scale=0.235, viewport=85 2 545 398]{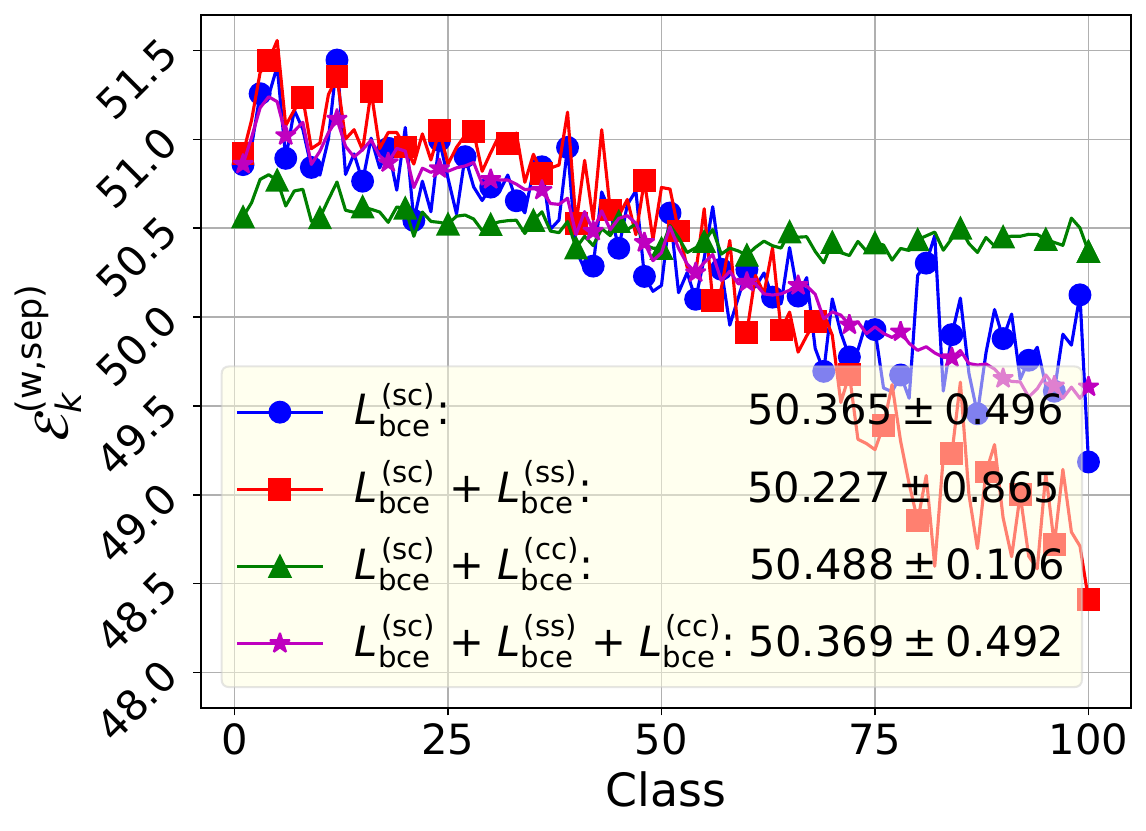}\\
    \caption{The intra-class compactness (top), inter-class separability (middle) of sample features, and separability (bottom) of classifier vectors on the training set of CIFAR100-LT (IF $=100$),
    with the model trained using different CE- (left) and BCE-based (right) methods.
    }
    \label{fig:CIFAR100_fu_comparsion_matrices}
    \vspace{-16pt}
\end{figure}

\begin{figure*}[tph]
    \centering
    \small
    \includegraphics[width=1.0\linewidth]{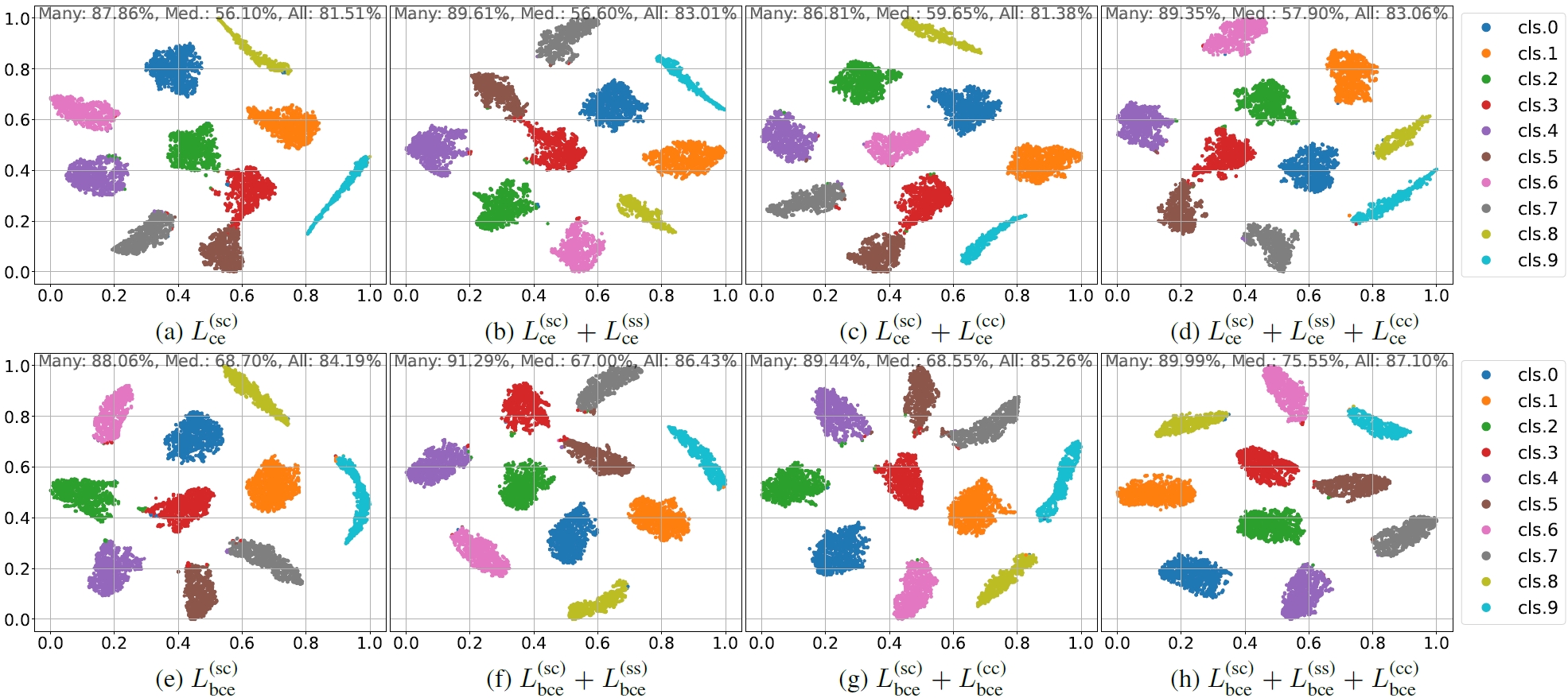}
    \vspace{-16pt}
    \caption{
        Feature distribution on the CIFAR10-LT test set with CE (top) and BCE (bottom) learning methods. Compared to CE methods, features extracted using BCE-based joint learning $L_{\text{bce}}^{\text{(sc)}}$ show improved intra-class compactness and inter-class separability. The contrastive learning $L_{\text{bce}}^{\text{(ss)}}$ and uniform learning $L_{\text{bce}}^{\text{(cc)}}$ further enhance these properties.
    } 
    \label{fig:CIFAR10_fu_comparsion_feature}
    \vspace{-8pt}
\end{figure*}

\begin{table}[t]
    \centering\small
    \arrayrulecolor{black}
        \setlength{\tabcolsep}{0.65mm}{
            \begin{tabular}{l|ccc|ccc}
            \arrayrulecolor{black}\hline
            \multirow{2}{*}{Methods} & \multicolumn{3}{c|}{CIFAR10-LT}                  & \multicolumn{3}{c}{CIFAR100-LT}                  \\
                                     & 100         & 50             & 10             & 100            & 50             & 10             \\
            \cline{1-6}\arrayrulecolor{black}\cline{7-7}
            CB-Focal\tiny{CVPR'19}                & 74.60          & 79.30          & 87.10          & 39.60          & 45.20          & 58.00          \\
            SSP\tiny{NeurIPS'20}     & 77.80          & 82.10          & 88.50          & 43.40          & 47.10          & 58.90          \\
            TSC\tiny{CVPR'22}            & 79.70          & 82.90          & 88.70          & 43.80          & 47.40          & 59.00          \\
            ETF+\small{DR}\tiny{NeurIPS'22}       & 76.50          & -              & 87.70          & 45.30          & -              & -              \\
            NC-DRW\tiny{AISTATS'23}     & 81.90          & -              & 89.80          & 48.60          & -              & 63.10          \\
            RIDE (3 exp.)\tiny{ICLR'21}  & -              & -              & -              & 48.60          & 51.40          & 59.80          \\
            BCL\tiny{CVPR'22}           & 84.50          & 87.20          & 91.10          & 51.90          & 56.40          & 64.60          \\
            NC.-cRT\tiny{AISTATS'23} & 82.60          & -              & 90.20          & 48.70          & -              & 63.60          \\
            ResLT+H2T~\tiny{AAAI'24}      & 81.77          & 84.99          & -              & 49.60          & 54.39          & -              \\
            Logit Adj.\tiny{ICLR'21} & 84.30          & 87.10          & 90.90          & 50.50          & 54.90          & 64.00          \\
            BCL+DODA\tiny{ICLR'24}         & -              & -              & -              & 51.00          & 53.60          & 62.70          \\
            RIDE+H2T\tiny{AAAI'24}     & -              & -              & -              & 51.38          & 55.54          & -              \\
            DiffuLT\tiny{NeurIPS'24}                         & 84.70          & 86.90          & 90.70          & 51.50          & 56.30          & 63.80          \\
            PaCo\tiny{ICCV'21}       & -              & -              & -              & 52.00          & 56.00          & 64.20          \\
            DiffuLT+RIDE\tiny{NeurIPS'24}           & 85.30          & 87.30          & 90.90          & 52.40          & 56.90          & 64.20          \\
            ProCo\tiny{TPAMI'24}  & 85.90          & 88.20          & 91.90          & 52.80          & 57.10          & 65.50          \\
            RBL\tiny{ICML'23}             & 84.70          & 87.60          & -              & 53.10          & 57.20          & -              \\
            DeiT-LT\tiny{CVPR'24}   & 87.50          & 89.80          & -              & 55.60          & 60.50          & -              \\
            GLMC\tiny{CVPR'23}                     & 88.50          & 91.04          & 94.87          & 57.97          & 63.78          & 73.40          \\
            GLMC + MN                               & \underline{89.58} & \underline{92.04} & \underline{94.87}          & \underline{58.41} & \underline{64.57} & \underline{74.28} \\
            BCE3S                               & \textbf{90.08} & \textbf{92.55} & \textbf{95.71} & \textbf{59.50} & \textbf{65.23} & \textbf{76.13} \\
            \arrayrulecolor{black}\cline{1-6}\arrayrulecolor{black}\cline{7-7}
            \end{tabular}
        }
    \arrayrulecolor{black}
    \caption{LTR on CIFAR10-LT and CIFAR100-LT, {IF} $=$ 100, 50, {and}~10. BCE3S consistently achieves the best results.}
    \label{table:benchmark_on_cifar}
\end{table}

{\textbf{Feature distribution in t-SNE.}}\quad To intuitively compare the feature properties between our BCE3S with CE-based methods,
for ResNet32 trained on CIFAR10-LT with IF $= 100$,
we visually show the feature distributions of the 10 classes on the test set in Fig. \ref{fig:CIFAR10_fu_comparsion_feature} using t-SNE.
As the figure shows, the feature clusters of class 3 and 5 (i.e., ``cat'' and ``dog'') of CE-based joint learning $L_{\text{ce}}^{\text{(sc)}}$ overlap with each other; though the intersection decreases with the addition of $L_{\text{ce}}^{\text{(ss)}}$ and $L_{\text{ce}}^{\text{(cc)}}$, the final feature clusters of the two classes are not completely separated, indicating unsatisfactory inter-class separability. Meanwhile, the features of the tail classes (the 8th and 9th classes) are spreading over relatively elongated regions, indicating unsatisfactory intra-class compactness. 
In contrast, for BCE, the ten feature clusters of $L_{\text{bce}}^{\text{(sc)}}$ are distributed in relatively independent regions, while with the addition of $L_{\text{bce}}^{\text{(ss)}}$ and $L_{\text{bce}}^{\text{(cc)}}$, the clusters of the tail classes becomes increasingly compact, indicating higher separability and compactness among the features.

\textbf{Parameter study.}\quad
BCE3S applied hyper-parameters $\lambda_{\text{ss}}$, $\lambda_{\text{cc}}$, and $\tau$ to balance the impact of three learning components, and we have conducted a comprehensive study using ResNet32 on CIFAR100-LT with IF $=$ 100.
The detailed experimental results for the parameter study can be found in the supplementary.

\subsection{Comparison with SOTA}
We compare our BCE3S with a series of SOTA LTR methods on CIFAR10-LT, CIFAR100-LT, ImageNet-LT, and iNaturalist2018 with different backbones.
To further explore the potential of BCE3S, we boost its performance by combining it with re-balancing techniques~\cite{2022_maxnorm,ren2020balanced_ms}. 
The detailed experimental settings and more results can be found in the supplementary.

On \textbf{CIFAR10-LT and CIFAR100-LT}, we adopted a similar training strategy used in MaxNorm (MN) of \citet{2022_maxnorm}.
As Table~\ref{table:benchmark_on_cifar} shows, our BCE3S achieves the best top-1 accuracy on both CIFAR10-LT and CIFAR100-LT with different IFs.
For example, on CIFAR100-LT, BCE3S achieves accuracies of $76.13\%$, $65.23\%$, and $59.50\%$ with the three IFs, 
which respectively surpass the previous best results by $1.85\%$, $0.66\%$, and $1.09\%$, being new SOTA.
On CIFAR10-LT dataset, BCE3S has also achieved SOTA results.
Those results highlight the potential of a model trained using BCE3S.

\begin{table}[h]
    \centering
    \small
    \arrayrulecolor{black}
    \ADLnullwidehline
    \setlength{\tabcolsep}{1.2mm}{
        \begin{tabular}{l|c|cccc}
        \arrayrulecolor{black}\hline
        \multirow{2}{*}{Methods} & \multirow{2}{*}{$\mathcal M$}    & \multicolumn{4}{c}{ImageNet-LT}          \\
                                &                              & \texttt{Many}       & \texttt{Med.}       & \texttt{Few}   & All    \\
        \cline{1-4}\arrayrulecolor{black}\cline{5-6}
        CB-Focal\tiny{CVPR'19} & \multirow{15}{*}{R50}  & 39.60      & 32.70      & 16.80 & 33.20  \\
        RISDA\tiny{AAAI'22}    &                        & -          & -          & -     & 49.30  \\
        LDAM\tiny{NeurIPS'19}  &                        & 60.40      & 46.90      & 30.70 & 49.80  \\
        KCL\tiny{ICLR'21}      &                        & 61.80      & 49.40      & 30.90 & 51.50  \\
        TSC\tiny{CVPR'22}      &                        & 63.50      & 49.40      & 30.40 & 52.40  \\
        GCL\tiny{CVPR'22}      &                        & 62.24      & 48.62      & 52.12 & 54.51  \\
        GCL+H2T\tiny{AAAI'24}  &                        & 62.36      & 48.75      & 52.15 & 54.62  \\
        BCL\tiny{CVPR'22}      &                        & 65.30      & 53.50      & 36.30 & 55.60  \\
        DiffuLT\tiny{NeurIPS'24}&                        & 63.20      & 55.40      & 39.20 & 56.20  \\
        BCL+DODA\tiny{ICLR'24}  &                        & 66.90      & 54.10      & 37.40 & 56.90  \\
        DiffuLT+RIDE\tiny{NeurIPS'24}&                        & 64.10      & \underline{55.80}      & \underline{39.90} & 56.90  \\
        DSCL\tiny{AAAI'24}      &                        & \textbf{68.50}      & 55.20      & 35.40 & 57.70  \\
        ProCo\tiny{TPAMI'24}    &                        & \underline{68.20} & 55.10      & 38.10 & \underline{57.80}  \\
        BCE3S                   &                        & 68.14      & \textbf{55.90} & \textbf{40.56} & \textbf{57.85}  \\
        \arrayrulecolor{black}\cline{1-4}\arrayrulecolor{black}\cline{5-6}
        \end{tabular}
    }
    \arrayrulecolor{black}
    \caption{LTR results on ImageNet-LT using ResNet50 (R50). BCE3S achieves the best results on the test set.}
    \label{table:benchmark_on_imagenet_lt_r50}
    \vspace{-16pt}
\end{table}

On \textbf{ImageNet-LT}, we evaluate BCE3S using ResNet50. As Table~\ref{table:benchmark_on_imagenet_lt_r50} shows, BCE3S achieves the best performance on \texttt{Medium}, and \texttt{Few}, with accuracy of 55.90\% and 40.56\%, and obtains the best overall accuracy of 57.85\%.

\begin{table}[h]
    \centering\small
    \arrayrulecolor{black}
    \setlength{\tabcolsep}{1.0mm}{
        \begin{tabular}{l|cccc}
        \arrayrulecolor{black}\hline
        \multirow{2}{*}{Methods} & \multicolumn{4}{c}{iNaturalist 2018} \\
                                & \texttt{Many}  & \texttt{Med.}  & \texttt{Few}   & All          \\
        \cline{1-2}\arrayrulecolor{black}\cline{3-5}
        KCL\tiny{ICLR'21}           & -     & -     & -     & 68.60        \\
        TSC\tiny{CVPR'22}           & 72.60 & 70.60 & 67.80 & 69.70        \\
        DR\tiny{ICLR'20}            & 72.88 & 71.15 & 69.24 & 70.49        \\
        GCL\tiny{CVPR'22}           & 66.43 & 71.66 & 72.47 & 71.47        \\
        GCL+H2T\tiny{AAAI'24}       & 67.74 & 71.92 & 72.22 & 71.62        \\
        BCL\tiny{CVPR'22}           & 69.40 & 72.40 & 71.80 & 71.80        \\
        DR+H2T\tiny{AAAI'24}        & 71.73 & 72.32 & 71.30 & 71.81        \\
        DSCL\tiny{AAAI'24}          & 74.20 & 72.90 & 70.30 & 72.00        \\
        RIDE\tiny{ICLR'21}          & 76.52 & 74.23 & 70.45 & 72.80        \\
        RIDE+H2T\tiny{AAAI'24}      & 75.69 & 74.22 & 71.36 & 73.11        \\
        PaCo (400 ep.)\tiny{ICCV'21}& 70.30 & 73.20 & 73.60 & 73.20        \\
        ProCo (90 ep.)\tiny{TPAMI'24}& -     & -     & -     & 73.50        \\
        BCL+DODA\tiny{ICLR'24}      & 71.20 & 73.20 & 73.40 & 73.70        \\
        BCE3S (180 ep.)             & \underline{77.16} & 74.45  & 72.78 & 73.99        \\\hdashline
        DeiT-LT(ViT-B, 1K ep.)\tiny{CVPR'24}    & 70.30 & 75.20 & \underline{76.20} & 75.10        \\
        ProCo(400 ep.)\tiny{TPAMI'24}   & 74.00 & \underline{76.00} & \textbf{76.80} & \underline{75.80} \\
        BCE3S  (400 ep.)                & \textbf{79.10} & \textbf{76.08} & 74.28 & \textbf{75.91} \\
        \arrayrulecolor{black}\cline{1-3}\arrayrulecolor{black}\cline{4-5}
        \end{tabular}
    }
    \arrayrulecolor{black}
    \caption{LTR results on iNaturalist2018 using ResNet50.}
    \label{table:benchmark_on_iNaturalist2018}
    \vspace{-16pt}
\end{table}

On \textbf{iNaturalist2018}, as Table~\ref{table:benchmark_on_iNaturalist2018} shows, we apply BCE3S to train ResNet50 using a strategy similar to that of \citet{du_2024_probabilistic_proco}.
When trained for 180 epochs, BCE3S achieves a competitive accuracy of $73.99\%$ on the overall test set, outperforming most previous methods. When extended to 400 epochs, BCE3S achieves the best performance on test subsets of \texttt{Many} and \texttt{Med.}, and get an optimal overall accuracy of $75.91\%$, surpassing ProCo and DeiT-LT, though DeiT-LT uses ViT-B trained for 1,000 epochs.

\section{Conclusion} \label{sec:conclusion}
For long-tailed recognition (LTR) on imbalanced datasets, this paper proposes the BCE-based tripartite synergistic learning method, i.e., BCE3S, by integrating the joint learning between sample features and classifier vectors, contrastive learning among the features, and uniform separability learning for the classifier vectors. 
Compared with CE-based methods, BCE3S suppresses the imbalance effect among the classifier vectors of the head and tail classes,
improves the models' focus on the feature learning in the tail classes, and enhances the intra-class compactness and inter-class separability of the features.
The extensive experiments demonstrate that our BCE3S achieves the optimal LTR performance on various long-tailed datasets.

\section*{Acknowledgements} \label{sec:acknowledgments}
The research was supported by National Natural Science Foundation of China under Grant No. 62576217 and 8226113862, Natural Science Foundation of Ningxia Province under Grant No. 2025AAC020002, Guangdong Provincial Key Laboratory under Grant No. 2023B1212060076, and Scientific Foundation for Youth Scholars of Shenzhen University under Grant No. 868-000001032180.

\bibliography{aaai2026}

\input{supplementary}

\end{document}

%% file: supplementary.tex
\clearpage

\setcounter{page}{1}
\appendix
\section*{Appendix A: Separability and compactness of classifier and sample features}\label{sec:three_metrics_supp}
\label{sec:three_metrics_supp}
Let $\mathcal{D}=\bigcup_{k=1}^K \mathcal{D}_k$
be a dataset from $K$ classes,
where $$\mathcal{D}_k = \Big\{\bm X_i^{(k)}\Big\}_{i=1}^{n_k}$$
contains $n_k$ samples from the $k$-th class.
In recognition task, for each sample $\bm X_i^{(k)}$, an encoder $\mathcal{M}(\cdot)$ first extracts its feature $\bm x_i^{(k)} = \mathcal M(\bm X_i^{(k)}) \in \mathbb R^d$,
then, a linear classifier $\mathcal C = \{(\bm w_j, b_j)\}_{j=1}^K$ converts it into $K$ metrics $\{\bm w_j^T \bm x_i^{(k)} + b_j\}_{j=1}^K$,
which are applied to predict the sample's label, $$\hat k = \arg\max_j\{\bm w_j^T \bm x_i^{(k)} + b_j\}_{j=1}^K.$$

In the ablation experiments, to compare the learned classifier and features, we define three metrics, separability $\mathcal{E}_k^{(\text{w,sep})}$ among classifier vectors, intra-class compactness $\mathcal{E}_k^{(\text{x,com})}$, and inter-class separability $\mathcal{E}_k^{(\text{x,sep})}$ among sample features, for any class $k$,
\begin{align}
    \mathcal{E}_k^{(\text{w,sep})}= & 
    \frac{1}{K-1}\sum_{j=1\atop j\neq k}^K\left(\frac{1-\cos(\hat{\boldsymbol{w}}_k,\hat{\boldsymbol{w}}_j)}{2}\right) \times 100\%\\
    \mathcal{E}_k^{(\text{x,com})} = & 
     \frac{1}{n_k^2-n_k} \sum_{i=1}^{n_k}\sum_{i'=1\atop i'\neq i}^{n_k}\left(\frac{\cos(\boldsymbol{x}_i^{(k)},\boldsymbol{x}_{i'}^{(k)})+1}{2}\right) \nonumber\\
    &\quad\times100\%, \\
    \mathcal{E}_k^{(\text{x,sep})}= &
    \frac{1}{n_kK-n_k} \sum_{i=1}^{n_k}\sum_{j=1\atop j\neq k}^K\left(\frac{1-\cos(\hat{\boldsymbol{x}}_i^{(k)},\bar{\boldsymbol{x}}^{(j)})}{2}\right) \nonumber\\
    &\quad\times 100\%,
\end{align}
where, for $\forall k$,
\begin{align}
    \hat{\bm w}_k &= \bm w_k - \bar{\bm w},\\
    \bar{\bm w}   &= \frac{1}{K}\sum_{j=1}^K\bm w_j,\\
    \bar{\bm x}^{(k)}   &= \frac{1}{n_k} \sum_{i=1}^{n_k} \bm{x}_i^{(k)},\\
    \hat{\bm{x}}_i^{(k)} &= \bm{x}_i^{(k)} - \bar{\bm{x}}^{(k)},
\end{align}
and $\bm x_i^{(k)}$ is the feature of the $i$-th sample from the $k$-th class.

\section*{Appendix B: Neural collapse}
The neural collapse was first found by \citet{papyan2020prevalence},
which refers to four properties about the sample features $\{\bm x_i^{(k)}\}$ and the classifier vectors $\{\bm w_k\}$
at the terminal phase of training. 
\begin{itemize}
\setlength{\topsep}{-1em}
\setlength{\itemsep}{0.1em}
\setlength{\leftmargin}{3em} %
\setlength{\parsep}{0em} %
\setlength{\labelsep}{0.5em} %
\setlength{\itemindent}{0.5em} %
\setlength{\listparindent}{0em} %
    \item {\bf NC1}, within-class variability collapse.
    Each feature $\bm x_i^{(k)}$ collapse to its class center $\bar{\bm x}^{(k)} = \frac{1}{n_k}\sum_{i'=1}^{n_k}\bm x_{i'}^{(k)}$, indicating the \emph{maximal intra-class compactness}
    \item {\bf NC2}, convergence to simplex equiangular tight frame.
    The set of class centers $\{\bar{\bm x}^{(k)}\}_{k=1}^K$  form a simplex equiangular tight frame (ETF),
    with equal and maximized cosine distance between every pair of feature means, i.e., the \emph{maximal  inter-class separability}.
    \item {\bf NC3}, convergence to self-duality.
    The class center $\bar{\bm x}^{(k)}$ is ideally aligned with the classifier vector $\bm w_k,\forall k\in[K]$.
    \item {\bf NC4}, simplification to nearest class center. The classifier is equivalent to a nearest class center decision.
\end{itemize}\vspace{-2pt}

The theory of neural collapse \cite{papyan2020prevalence,zhu2021geometric,fang_2021_exploring_nc} indicates that the centered classifier vectors form an ETF framework at terminal training, i.e.,
\begin{align}
    \label{eq:ETF_condition}
    &\cos(\hat{\bm w}_k, \hat{\bm w}_{k'}) = \frac{K}{K-1}\delta_{k,k'} - \frac{1}{K-1},~~\text{and}\\
    &\|\bm w_k\| = \|\bm w_{k'}\|,~~\forall k,k',
\end{align}
where $\delta_{k,k'}$ denotes the Kronecker delta function, which equals $1$ when $k=k'$ and $0$ otherwise.

From Fig. \ref{fig:CIFAR100_fu_comparsion_matrices} and Fig. \ref{supp_fig:CIFAR100_fu_comparsion_matrices_train}, on CIFAR100-LT one can find the separability of any classifier vectors is close to $50.50$ which is approximately equal to $\frac{1+\frac{1}{K-1}}{2} \times 100 = \frac{500}{99}$.

\section*{Appendix C: Tripartite synergistic learning}
In our BCE3S, we apply symbols ``sc'', ``ss'', and ``cc'' to indicate the joint learning, contrastive learning, and uniform learning, as they measures the metrics of pairs of sample-to-class, sample-to-sample, and class-to-class, respectively.
Besides BCE3S, one can also design CE-based tripartite synergistic learning which also consists of three components: CE-based joint learning $L_{\text{ce}}^{\text{(sc)}}$, CE-based contrastive learning $L_{\text{ce}}^{\text{(ss)}}$, and CE-based uniform learning $L_{\text{ce}}^{\text{(cc)}}$.

\textbf{CE-based joint learning.}\quad
During training, for any sample $\bm X^{(k)}$ from the $k$-th class, the model $\mathcal M$ extracts its feature as $\bm x^{(k)} = \mathcal M(\bm X^{(k)})$, and the classifier $\mathcal C = \{(\bm w_j, b_j)\}_{j=1}^K$ converts the feature into $K$ metrics $\{\bm w_j^T \bm x + b_j\}_{j=1}^K$, then Softmax is used to compute class probabilities $\big\{ \frac{\exp(\bm{w}_j^T \bm{x}^{(k)} + b_j)}{\sum_{\ell=1}^K \exp(\bm{w}_\ell^T \bm{x}^{(k)} + b_\ell)} \big\}_{j=1}^K$ and cross-entropy is used to compute the loss for joint learning between sample feature and classifier,
\begin{align}
    \label{eq:sample_to_class_CE_tri}
        L_{\text{ce}}^{\text{(sc)}}(\bm x^{(k)}) = -\log\Big(\frac{\exp{({\bm w}_k^T \bm x^{(k)} + b_k)}}{\sum_{j=1}^{K}\exp{({\bm w}_j^T \bm x^{(k)} + b_j)}}\Big),
\end{align}
where $\|{\bm w}_j\| = 1$ for $\forall j$.

\textbf{CE-based contrastive learning.}\quad
Following SimCLR~\cite{chen_simple_2020} and GLMtC~\cite{du2023glmc}, we implement contrastive learning in a projection space. For any sample feature $\bm x^{(k)}$ from class $k$, we transform the sample feature into $\bm z^{(k)} = \mathcal P(\bm x^{(k)})\in \mathbb R^{d'}$ using a non-linear projector $\mathcal P$, and
\begin{align}
    \label{eq:sample_to_sample_CE_tri}
    L_{\text{ce}}^{\text{(ss)}}(\bm x^{(k)}) = -\log\Big(\frac{\exp\big(\frac{1}{\tau}\cos(\bm z^{(k)},\bm z_*^{(k)})\big)}
                {\sum_{j=1}^{K}\exp\big(\frac{1}{\tau}\cos(\bm z^{(j)},\bm z_*^{(k)})\big)}\Big),
\end{align}
where $\big\{\bm z_*^{(j)} = \mathcal P(\bm x_*^{(j)})\big\}_{j=1}^K$ are projections of $K$ sample features $\big\{\bm x_*^{(j)}\big\}_{j=1}^K$ stored in a memory bank, and $\tau$ is a temperature factor.
Similar to SimCLR~\cite{chen_simple_2020}, the non-linear projector $\mathcal P$ consists of a two-layer MLP with a non-linear activation function. The CE-based contrastive learning is similar to that in GLMC~\cite{du2023glmc}.

\textbf{CE-based uniform learning.}\quad
Softmax and cross-entropy can also be used to design uniform learning for the classifier,
\begin{align}
    \label{eq:class_to_class_CE_tri}
        L_{\text{ce}}^{\text{(sc)}}(\bm w_k) = -\log\Big(\frac{\exp{({\bm w}_k^T {\bm w}_k)}}{\sum_{j=1}^{K}\exp{({\bm w}_k^T {\bm w}_j)}}\Big),
\end{align}
where the numerator in the Softmax is natural constant $\e$.

\textbf{CE3S.}\quad
During the training, a batch of samples $[\bm X_i^{(k_i)}]_{i=1}^B$ are fed into the model $\mathcal M$, and their features $[\bm x_i^{(k_i)}]_{i=1}^B$ are extracted, where $B$ is the batch size. The model training can be driven by the CE3S loss,
\begin{align}
    \label{eq:ce_tri_learning_loss}
    L^{(\text{tri})}_{\text{ce}}\big([\bm x_i^{(k_i)}]\big) &= \frac{1}{B}\sum_{i=1}^B L^{(\text{sc})}_{\text{ce}}(\bm x_i^{(k_i)})
                                   + \frac{\lambda_{\text{ss}}}{B}\sum_{i=1}^B L^{(\text{ss})}_{\text{ce}}(\bm x_i^{(k_i)}) \nonumber\\
                                  &+ \frac{\lambda_{\text{cc}}}{K}\sum_{k=1}^K L^{(\text{cc})}_{\text{ce}}(\bm w_{k}).
\end{align}

If the models were trained using a two-stage training pipeline, we will take the whole CE3S in the first stage and only fine-tune the classifier vectors by using a class-balanced CE-based joint learning, similar to that in the work of \citet{cui2019class},
\begin{align}
    L_{\text{ce-cb}}^{(\text{sc})}\big([\bm x_i^{(k_i)}]\big) = \frac{1}{B}\sum_{i=1}^B\frac{1-\beta}{1-\beta^{n_{k_i}}} L_{\text{ce}}^{\text{(sc)}}(\bm x_i^{(k_i)}),
\end{align}
where $\beta$ is a re-weighting parameter. In our experiments with CE3S or BCE3S, we set $\beta=0.9999$.

\section*{Appendix D: CE vs. BCE in LTR}
In the main body of paper, we have in-depth analyzed and compared BCE-based and CE-based joint learning with respect to feature learning.
We here compare their contrastive learning and uniform learning in terms of the learning and features and classifier.

\textbf{Contrastive learning for feature learning.}\quad
With BCE- and CE-based contrastive learning $L_{\text{bce}}^{\text{(ss)}}$ and $L_{\text{ce}}^{\text{(ss)}}$, the sample features are also updated in the projection space through their gradients,
\begin{equation}
    \bm z^{(k)} \leftarrow \bm z^{(k)} - \eta \frac{\partial L^{\text{(ss)}}_{\mu}\big(\bm z^{(k)}\big)}{\partial \bm z^{(k)}},
\end{equation}
where $\mu \in \{\text{ce}, \text{bce}\}$ denotes CE- or BCE-based contrastive learning, and $\eta$ is the learning rate.
The gradients $\frac{\partial L_{\mu}^{(\text{ss})}(\bm z^{(k)})}{\partial \bm z^{(k)}}$ have the similar form in CE or BCE,
\begin{align}
    \frac{\partial L_{\mu}^{(\text{ss})}(\bm z^{(k)})}{\partial \bm z^{(k)}} = &\underbrace{-\frac{1}{\tau}\big(1- \mathtt{Act}_{\mu}(\cos(\bm z_*^{(k)},\bm z^{(k)}))\big)\bm z_*^{(k)}}_{\text{pulling}} \nonumber\\
     + &~\sum_{j=1 \atop j\neq k}^K\underbrace{\frac{1}{\tau}\mathtt{Act}_{\mu}(\cos(\bm z_*^{(j)},\bm z^{(k)})) \bm z_*^{(j)}}_{\text{repelling}},
\end{align}
where, for $\forall j$,
\begin{align}
    &\mathtt{Act}_{\text{ce}}(\cos(\bm z_*^{(j)},\bm z^{(k)})) = \frac{\exp(\cos(\bm z_*^{(j)},\bm z^{(k)}))}{\sum_{\ell=1}^K \exp(\cos(\bm z_*^{(\ell)},\bm z^{(k)}))},\\
    &\mathtt{Act}_{\text{bce}}(\cos(\bm z_*^{(j)},\bm z^{(k)})) = \frac{1}{1+\e^{-\cos(\bm z_*^{(j)},\bm z^{(k)})}}.
\end{align}

In the memory bank, the projected feature $\bm z_*^{(k)}$ is updated by the last feature $\bm x^{(k_i=k)}_i$ from the same class in the online batch;
if there is no data from the same class $k$, the projected feature $\bm z_*^{(k)}$ is not updated in this iteration using the batch.
As the probability of samples from different classes in the long-tailed dataset appearing in the batch is different, so the learning degree of the projected features $\big\{\bm z_*^{(k)}\big\}_{k=1}^K$ in the memory bank is not the same, resulting in strong imbalance.
The imbalanced $K$ projected features are coupled in the denominator of Softmax $\mathtt{Act}_{\text{ce}}$ by CE-based contrastive learning, while they are decoupled via Sigmoid $\mathtt{Act}_{\text{bce}}$ by BCE-based contrastive learning. Therefore, $L_{\text{bce}}^{\text{(ss)}}$ will achieve more balance features than $L_{\text{ce}}^{\text{(ss)}}$.

\textbf{Uniform learning for classifier vectors.}\quad
In the training with the uniform learning, $L_{\text{ce}}^{\text{(cc)}}$ and $L_{\text{bce}}^{\text{(cc)}}$, the classifier vectors are updated via their gradients,
\begin{align}
    {\bm w}_k \leftarrow {\bm w}_k - \eta \frac{\partial L^{\text{(cc)}}_{\mu}\big({\bm w}_k\big)}{\partial {\bm w}_k},~~\forall\,k,
\end{align}
where $\mu \in \{\text{ce}, \text{bce}\}$ denotes CE- or BCE-based uniform learning, and $\eta$ is the learning rate.
The gradients $\frac{\partial L_{\mu}^{(\text{cc})}(\bm w_k)}{\partial {\bm w}_k}$ have the similar form in CE or BCE,
\begin{align}
    \underbrace{-\big(1- \mathtt{Act}_{\mu}({\bm w}_k ^T {\bm w}_k)\big){\bm w}_k}_{\text{pulling}} + \sum_{j=1 \atop j\neq k}^K\underbrace{ \mathtt{Act}_{\mu}({\bm w}_j ^T {\bm w}_k) {\bm w}_j}_{\text{repelling}},
\end{align}
where, for $\forall j$,
\begin{align}
    \mathtt{Act}_{\text{ce}}({\bm w}_j ^T {\bm w}_k)  &= \frac{\exp({\bm w}_j^T  {\bm w}_k)}{\sum_{\ell=1}^K \exp({\bm w}_\ell^T  {\bm w}_k)},\\
    \mathtt{Act}_{\text{bce}}({\bm w}_j ^T {\bm w}_k) &= \frac{1}{1+\e^{-{\bm w}_j ^T {\bm w}_k}}.
\end{align}

In updating classifier vectors, both $L_{\text{ce}}^{\text{(cc)}}$ and $L_{\text{bce}}^{\text{(cc)}}$ utilize exponential similarities between a classifier vector and all $K$ classifier vectors to compute one pulling term and $K-1$ repelling terms. For CE-based uniform learning ($L_{\text{ce}}^{\text{(cc)}}$), the computation of each pulling or repelling term involves $K$ imbalanced exponential similarities coupled in the Softmax denominator ($\mathtt{Act}_{\text{ce}}$), thereby injecting the imbalance effect into the uniform learning process.
In contrast, BCE decouples the metrics between the $K$ classifier vectors, employing only a single similarity to compute each pulling or repelling term through the Sigmoid function $\mathtt{Act}_{\text{bce}}$. Consequently, $L_{\text{bce}}^{\text{(cc)}}$ achieves more uniform separability across the $K$ classifier vectors.

\section*{Appendix E: Datasets and training details}\label{sec:supp_train_detail}
We evaluate the proposed BCE-based tripartite synergistic learning (BCE3S) on four popular LTR datasets,
including CIFAR10-LT \cite{yue2016deep}, CIFAR100-LT \cite{yue2016deep}, ImageNet-LT \cite{liu2022open_imagenet_lt}, and iNaturalist2018 \cite{van2018inaturalist}.

\textbf{CIFAR10-LT and CIFAR100-LT.}\quad
Following the work of \citet{yue2016deep}, the long-tailed CIFAR10 and CIFAR100 are produced by sampling the training samples of the original datasets, using an exponential decay imbalance mode across classes, and the long-tailed datasets are referred to as CIFAR10-LT and CIFAR100-LT, respectively.
For each of them, we produced their three different variants using three different imbalance factors (IF), 100, 50, and 10.

\begin{figure}[t]
    \centering
    \subfloat[Sample distribution in the training set of CIFAR10-LT with IF = 100.]
        {\label{supp_fig:cifar10_lt_description}\includegraphics*[scale=0.18, viewport=-140 45 1140 530]{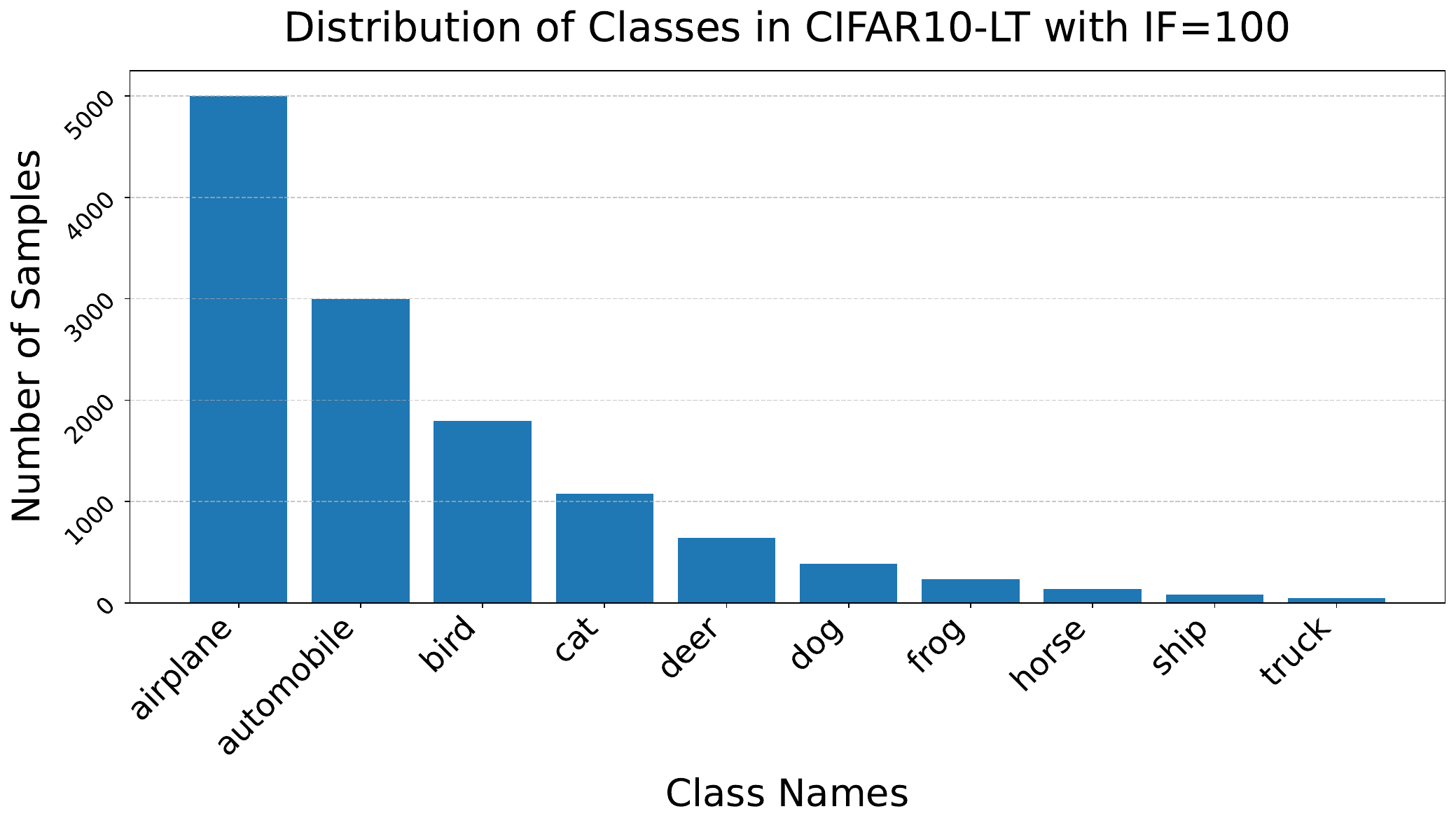}}\\
    \subfloat[Sample distribution in the training set of CIFAR100-LT with IF = 100.]
        {\label{supp_fig:cifar100_lt_description}\includegraphics*[scale=0.18, viewport=0 40 1280 530]{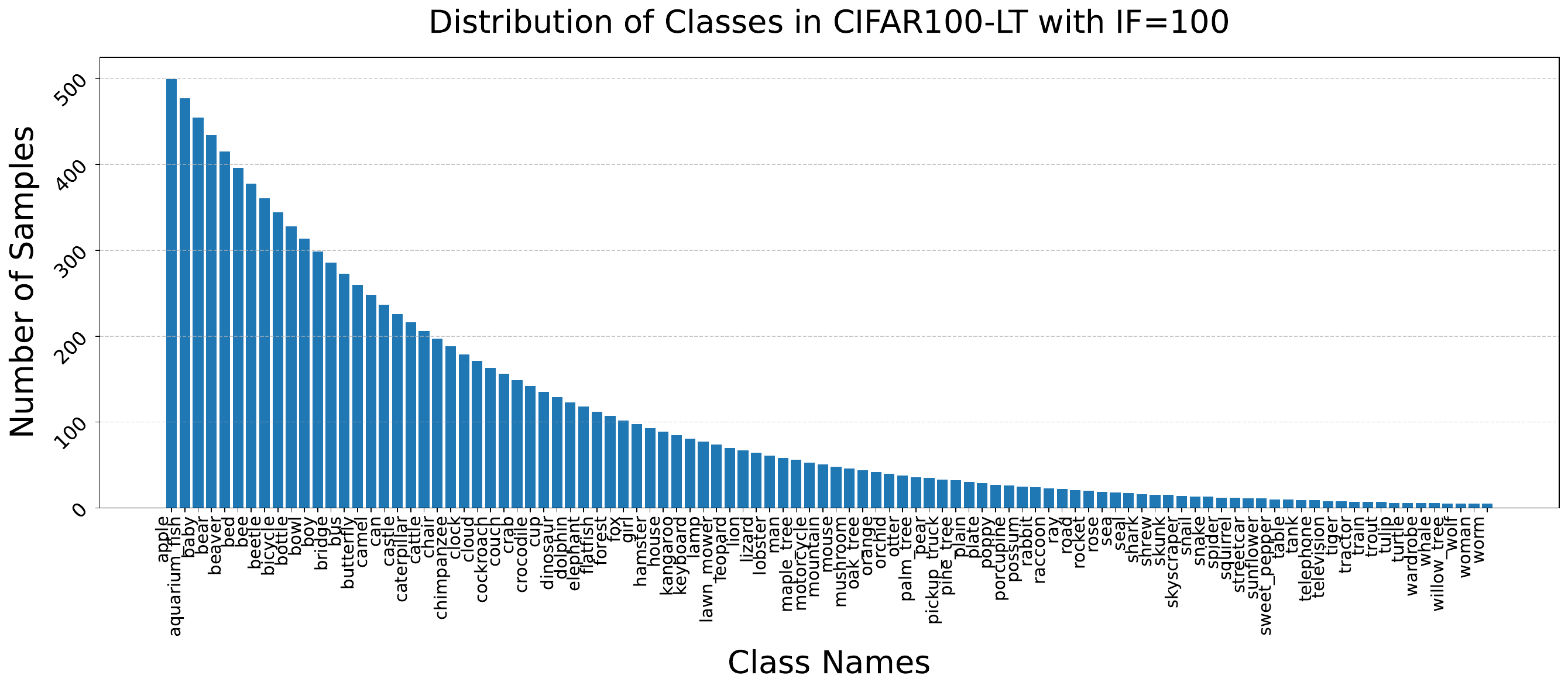}}\\
    \subfloat[Sample distribution in the training set of ImageNet-LT.]
        {\label{supp_fig:imagenet_lt_description}\includegraphics*[scale=0.18, viewport=0 0 1100 388]{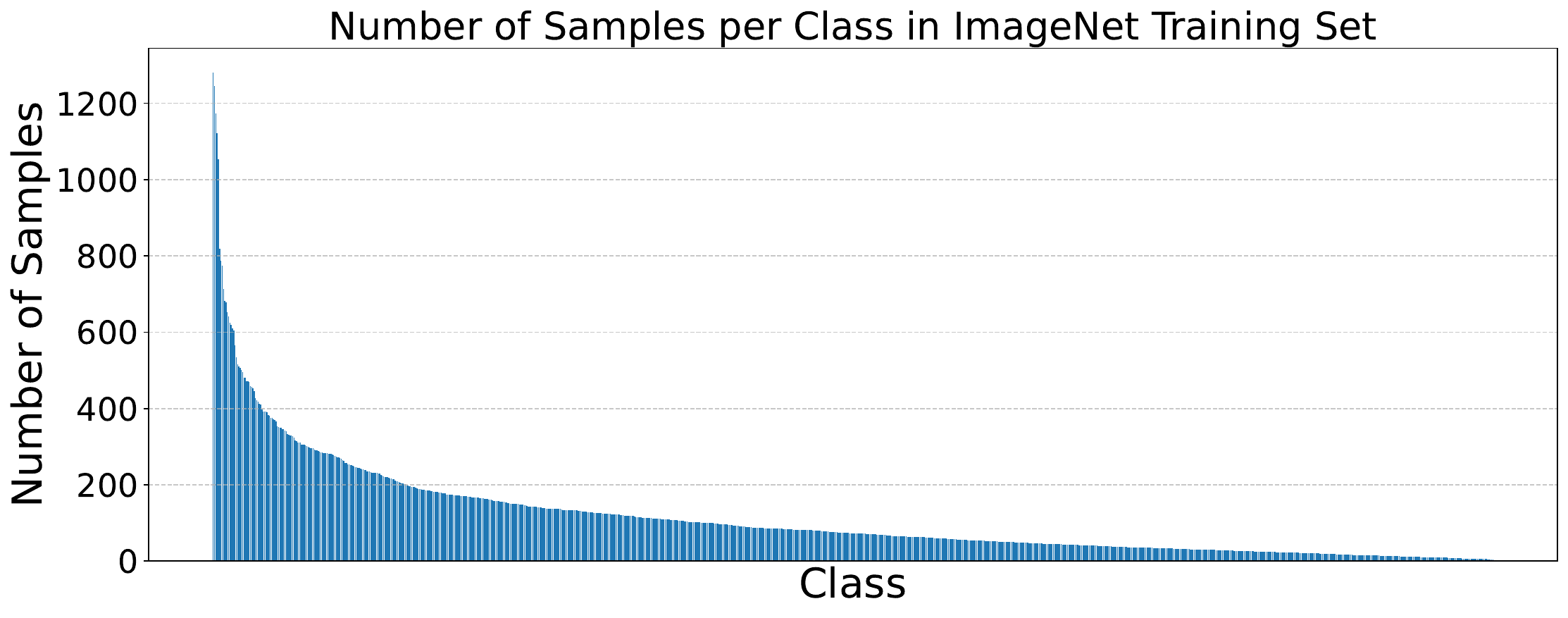}}\\
    \subfloat[Sample distribution in the training set of iNaturalist2018.]
        {\label{supp_fig:inaturalist18_description}\includegraphics*[scale=0.18, viewport=0 0 1280 388]{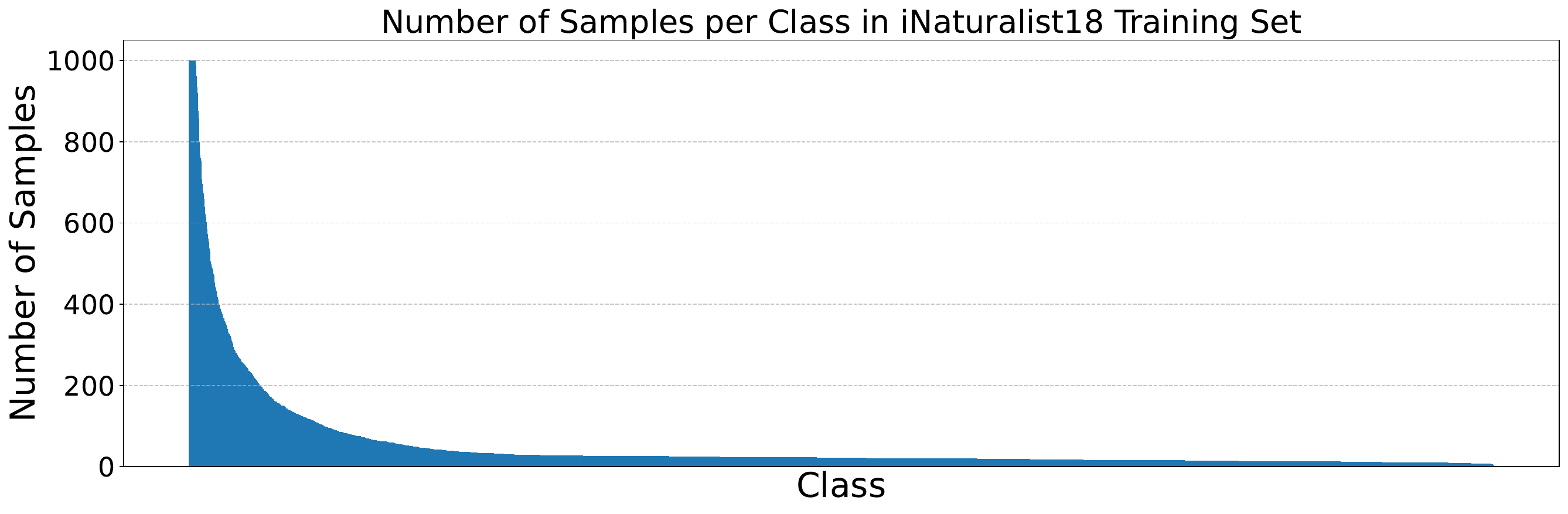}}
        \caption{Dataset descriptions for different long-tailed datasets.}
    \label{supp_fig:all_datasets_description}
\end{figure}

Fig.~\ref{supp_fig:cifar10_lt_description} illustrates the sample numbers of the ten classes in training set of CIFAR10-LT with IF $= 100$.
The dataset comprises samples from 10 classes, i.e., airplane, automobile, bird, cat, deer, dog, frog, horse, ship, and truck.
The first head class (airplane) contains 5,000 samples, while the last tail class (truck) has only 50 samples.
The smallest class of CIFAR10-LT contains 50 samples at least, thereby no \texttt{few} subset exists.

Fig.~\ref{supp_fig:cifar100_lt_description} illustrates the distribution of the sample numbers of the 100 classes in the training set of CIFAR100-LT with IF of 100.
The dataset comprises 100 distinct classes, with sample numbers ranging from 500 to merely 5, following a smooth exponential decay pattern across all classes.

\textbf{ImageNet-LT} \cite{liu2022open_imagenet_lt} is produced by sampling a subset of ImageNet \cite{russakovsky2015imagenet} using the Pareto distribution with a power factor of $\alpha_p=6$, which comprises 115,846 images from 1,000 classes and the sample number from the head class to the tail class decreases from 1,280 to 5, resulting in IF $=256$.
Fig.~\ref{supp_fig:imagenet_lt_description} illustrates the sample distribution in different classes in its training set; we sort the classes in descending order according to the sample number.

\textbf{iNaturalist18.}\quad
This large-scale dataset, collected from the iNaturalist species identification platform, encompasses 437,513 images across 8,142 species classes, which exhibits a natural long-tailed distribution, with sample numbers ranging from 1,000 for the most common species to merely 2 for the rarest ones, resulting in an IF of 500.
This extreme class imbalance authentically reflects real-world bio-diversity observation patterns, where certain species are frequently encountered while others are exceptionally rare. Fig.~\ref{supp_fig:inaturalist18_description} illustrates the sample distribution in its training set.

\textbf{Training Details.}\quad
We take ResNets as the classification models,
and train them on each of the imbalanced datasets from scratch.
For each training, we employ SGD optimizer with a momentum of $0.9$ and decay learning rate with cosine scheduler.
For the training on CIFAR10-LT and CIFAR100-LT, we adopt a customized ResNet32, and set the initial learning rate as $0.01$ and weight decay rate as $5$e$-3$, which are the same with that used in MaxNorm \cite{2022_maxnorm}.
For a fair comparison, we try to reproduce the results in MaxNorm \cite{2022_maxnorm}, and we got the recognition accuracy reported in the paper using the official code by training the ResNet32 with 320 epochs. Therefore, we evaluate our BCE3S with a training of 320 epochs.

On ImageNet-LT, following DSCL~\cite{xuan2024_aaai_dscl}, we train ResNet50 and ResNeXt50 for 200 epochs with the initial learning rate of $0.05$, weight decay rate of $5$e$-4$, and batch size of $128$.
On iNaturalist2018, following ProCo~\cite{du_2024_probabilistic_proco}, we train ResNet50 for 180, 400 epochs, respectively, with initial learning rate of $0.1$, weight decay of $1$e$-4$, and batch size of $256$.

For each model trained on the imbalanced training set, it is evaluated on the corresponding balanced test set.
Following~\cite{Kang_2020_Decoupling,2022_maxnorm,liu2022open, du_2024_probabilistic_proco}, besides the total recognition accuracy on the whole test set, we also report the accuracy on three kinds of subsets, i.e., \texttt{Many}, \texttt{Medium}, and \texttt{Few}, while the numbers of corresponding training samples in each class are larger than $100$, varying from $20$ to $100$, and less than $20$, respectively.

\textbf{BCE3S with re-balancing methods.}\quad
Similar to previous approaches~\cite{Kang_2020_Decoupling, 2022_maxnorm, xuan2024_aaai_dscl}, BCE3S's performance in long-tailed recognition can be further improved through a two-stage training strategy. In the first stage, the model is trained using the complete BCE3S method. In the second stage, we fix the feature extractor parameters and only fine-tune the classifier using a class-balanced binary cross-entropy loss \cite{cui2019class}, defined as:
\begin{align}
L_{\text{bce-cb}}^{(\text{sc})}\big([\bm x_i^{(k_i)}]\big) = \frac{1}{B}\sum_{i=1}^B\frac{1-\beta}{1-\beta^{n_{k_i}}} L_{\text{bce}}^{\text{(sc)}}(\bm x_i^{(k_i)}),
\end{align}
where $\beta$ is a hyperparameter that controls the degree of re-weighting based on class frequency. BCE3S's performance can be boosted by existing re-balancing methods, which demonstrates the compatibility and potential of BCE3S.

\section*{Appendix F: Normalized sample feature and classifier}
Similar to the previous work~\cite{Kang_2020_Decoupling}, we adopt normalized classifier in the BCE-based joint learning $L_{\text{bce}}^{\text{(sc)}}$.
We here compare the different normalization modes on the sample feature or classifier vectors.
We train ResNet32 using $L_{\text{bce}}^{\text{(sc)}}$ on CIFAR100-LT with IF = 100.
Table~\ref{table:normalzie_comparasion} shows the results. One can find that if normalization is not applied or applied simultaneously on sample features and classifiers, the final accuracy of the model is similar, 49.89\% and 49.93\%, respectively. When the normalization is only applied on the sample features, the model training fails. 
If normalization is applied on the classifier, the model accuracy reaches a maximum of 51.56\%. 
Therefore, in BCE3S, we only normalize the classifier vector for the joint learning.

\begin{table}[th]
    \centering\small
    \arrayrulecolor{black}
    \begin{tabular}{cc|cccc}
    \hline
    \multicolumn{2}{c|}{Normalization} & \multirow{2}{*}{\texttt{Many}} & \multirow{2}{*}{\texttt{Med.}} & \multirow{2}{*}{\texttt{Few}} & \multirow{2}{*}{All}  \\
    \arrayrulecolor{black}\cline{1-2}
    feature & Classifier                   &                       &                        &                      &                       \\
    \hline
    ~       & ~                        & 81.20                 & 52.51                  & 10.30                & 49.89                 \\
    \checkmark       & ~               & 2.86                  & 0.00                   & 0.00                 & 1.00                  \\
    ~       & \checkmark               & 81.83                 & 52.20                  & 15.50                & 51.56                 \\
    \checkmark       & \checkmark      & 79.60                 & 49.11                  & 16.27                & 49.93                 \\
    \arrayrulecolor{black}\cline{1-4}\arrayrulecolor{black}\cline{5-6}
    \end{tabular}
    \arrayrulecolor{black}
    \caption{Comparison of normalized sample features and classifier vectors. All experiments employ BCE-based joint learning, $L_{\text{bce}}^{\text{(sc)}}$ with $r=1.0$ on CIFAR100-LT dataset with IF = 100.}
    \label{table:normalzie_comparasion}
\end{table}

\begin{figure}[t]
    \centering
    \small
    \includegraphics*[scale=0.205, viewport=8 0 565 455]{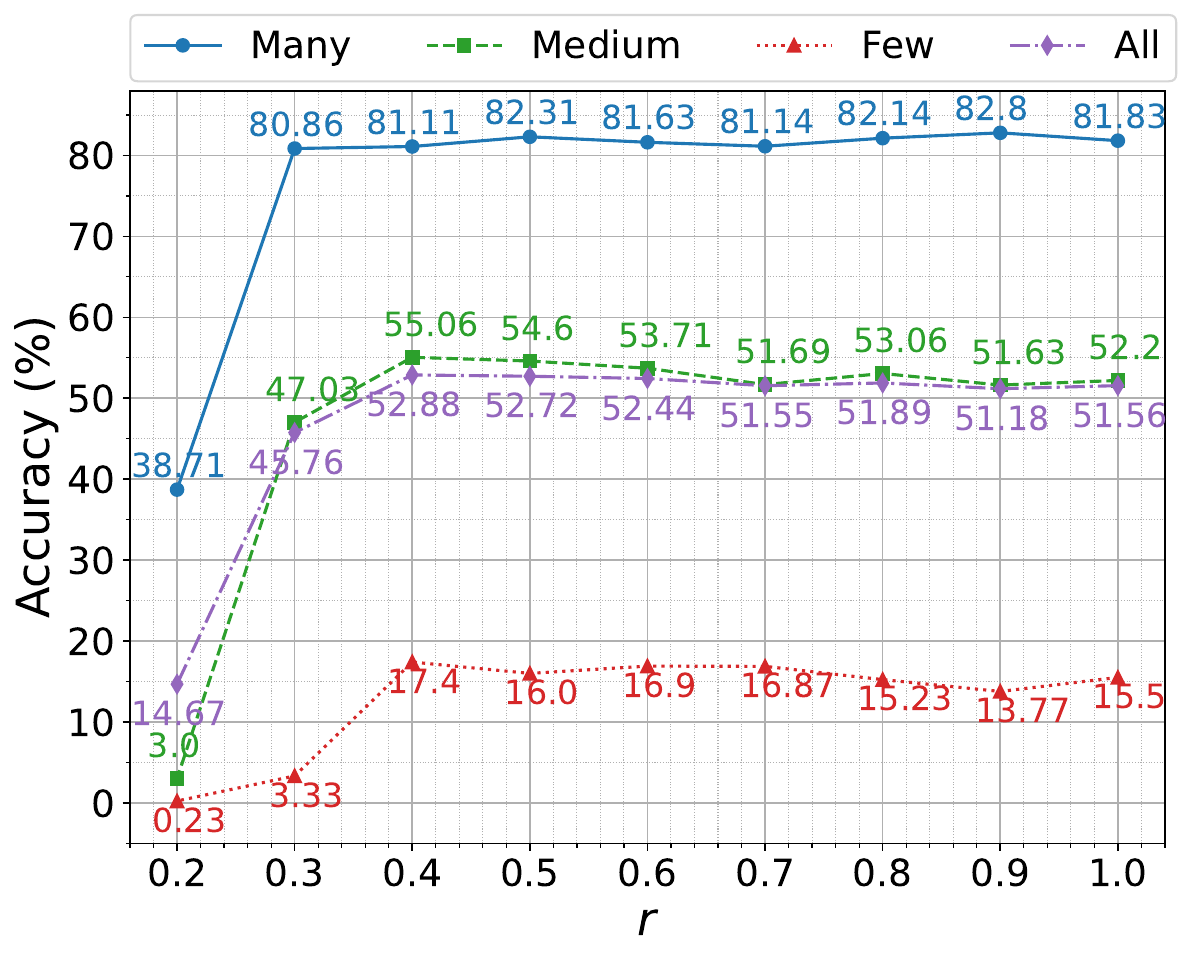}
    \includegraphics*[scale=0.215, viewport=8 0 565 420]{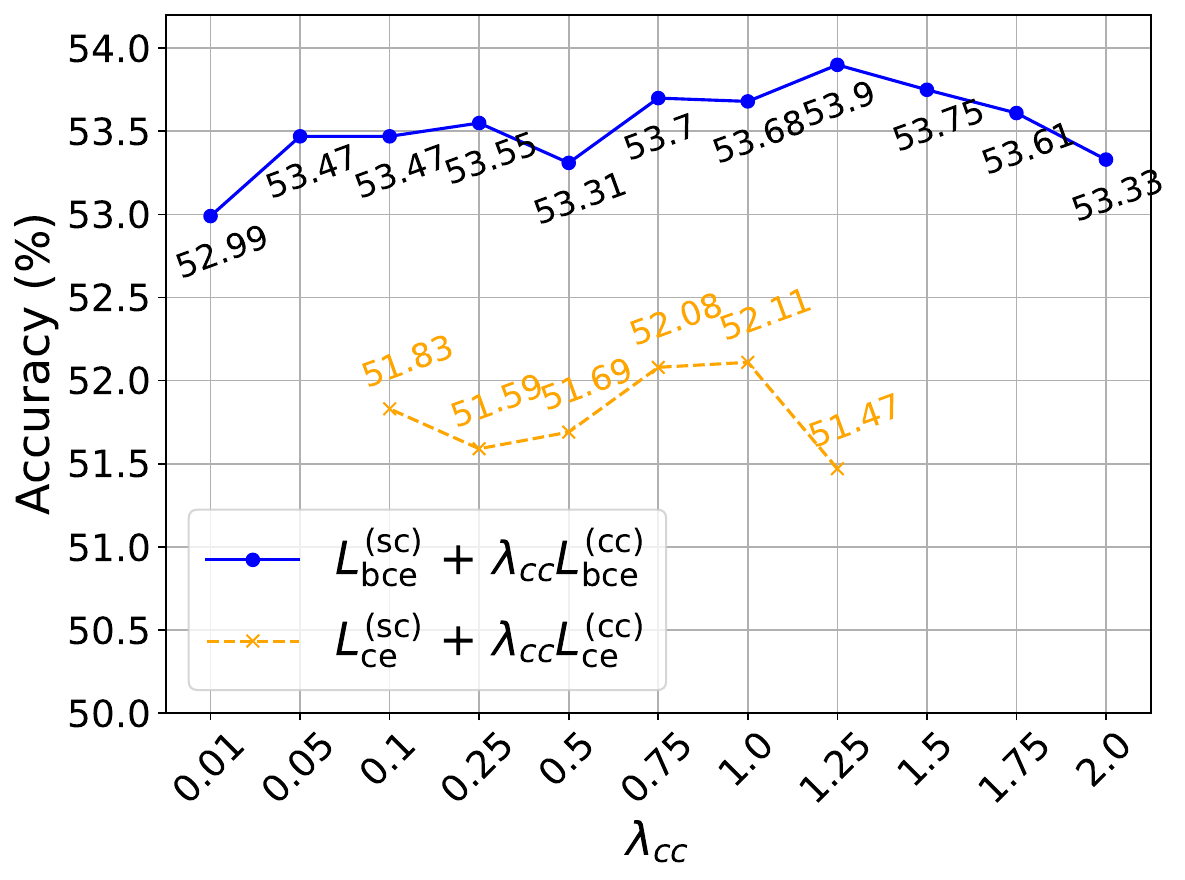}\\
    \includegraphics*[scale=0.235, viewport=8 5 770 272]{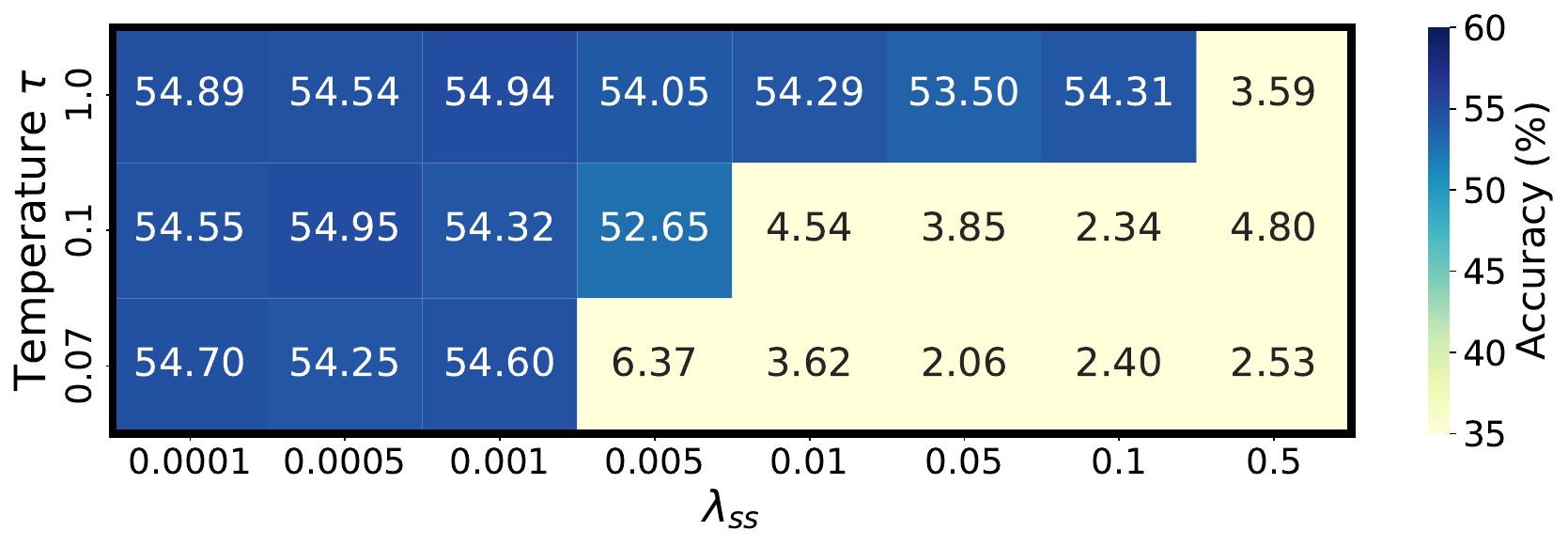}\vspace{-10pt}
    \caption{Parameter study for $L_{\text{bce}}^{\text{(sc)}}$ (top left), $L_{\text{bce}}^{\text{(ss)}}$ (bottom) and $L_{\text{bce}}^{\text{(cc)}}$ (top right), on CIFAR100-LT with IF $=100$.}
    \label{fig:parameters_study}\vspace{-16pt}
\end{figure}

\begin{figure*}[t]
    \centering
    \includegraphics*[scale=0.33, viewport=-2 25 534 388]{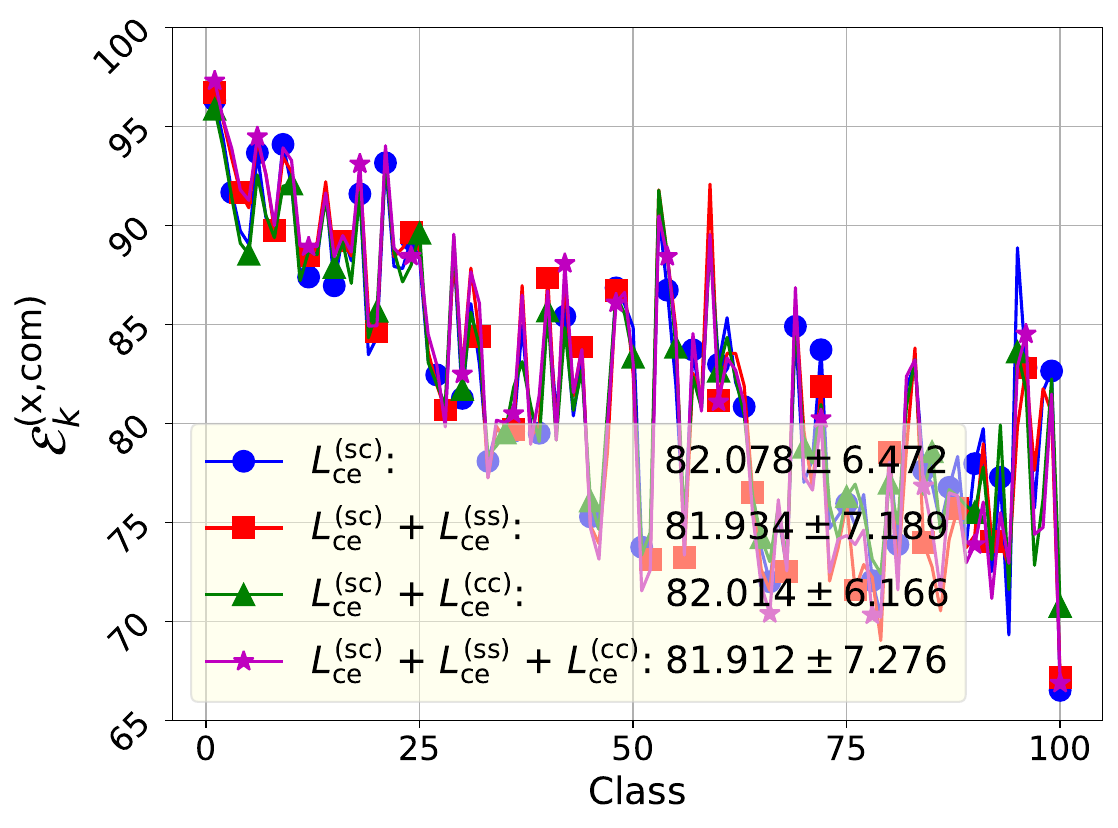}\hspace{-2.6pt}
    \includegraphics*[scale=0.33, viewport=80 25 535 388]{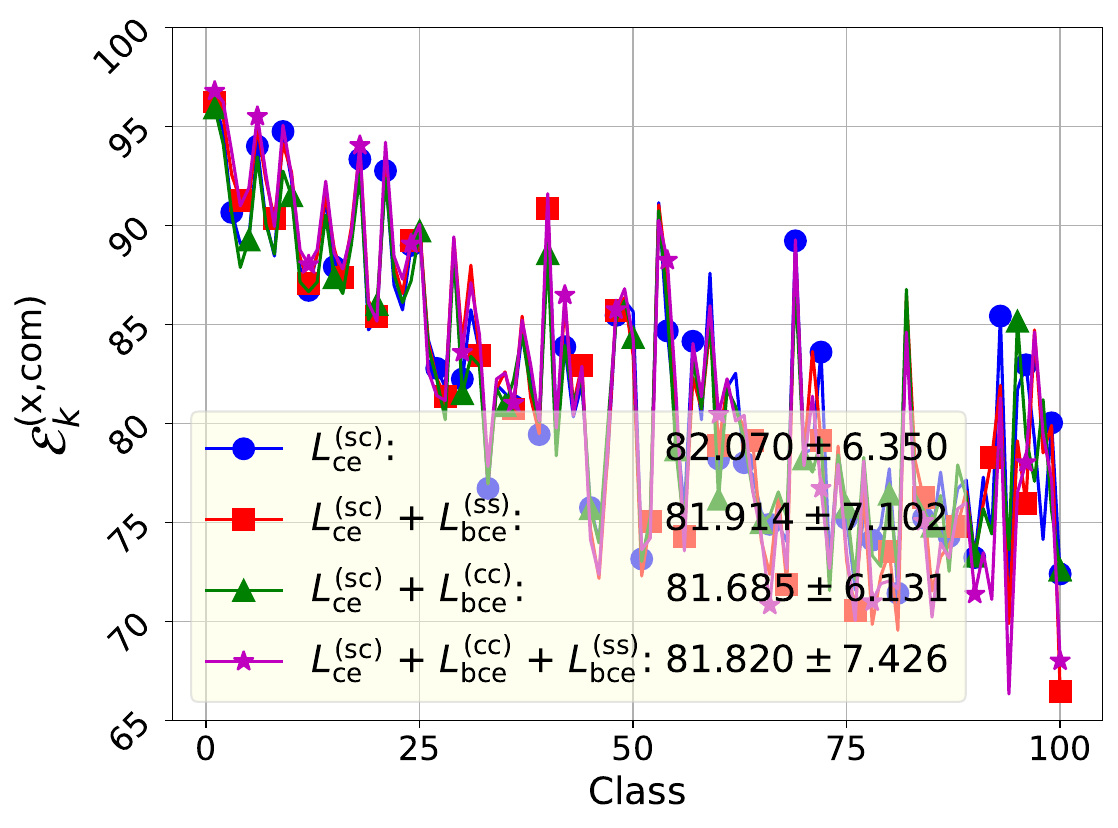}\hspace{-2pt}
    \includegraphics*[scale=0.33, viewport=80 25 535 388]{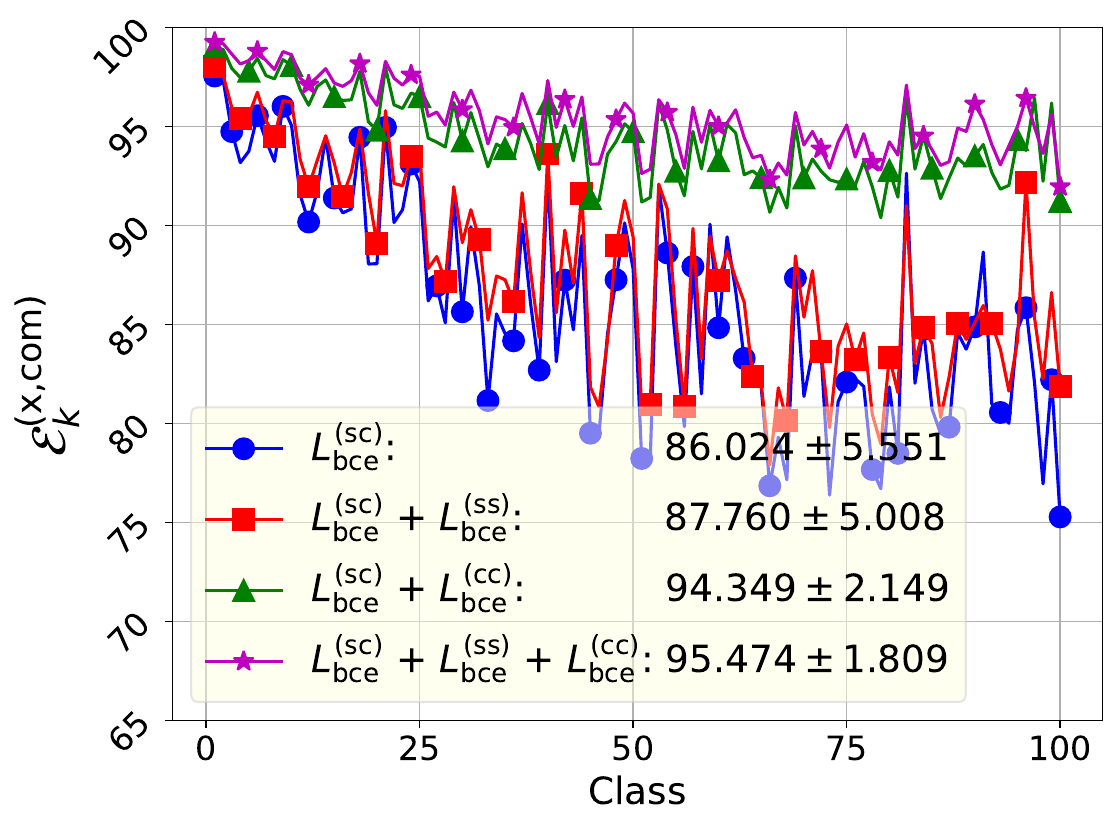}\\

    \includegraphics*[scale=0.33, viewport=8  25 534 388] {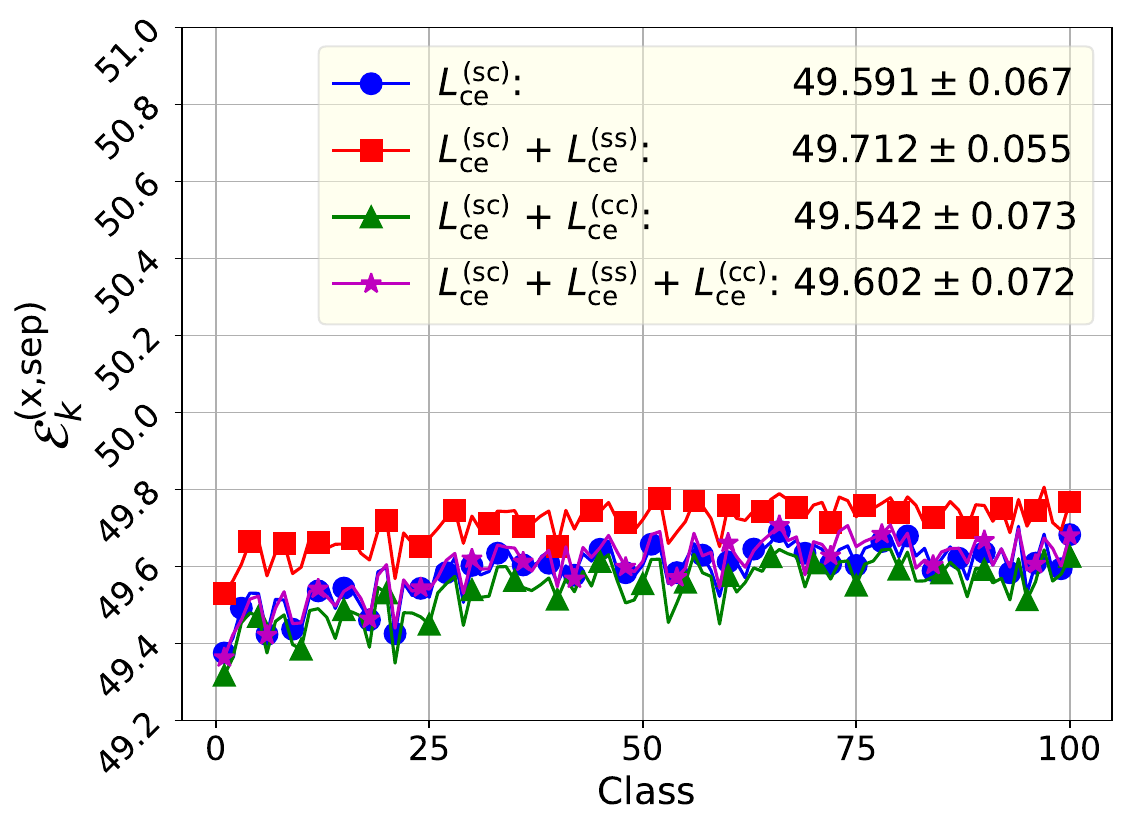}\hspace{-2pt}
    \includegraphics*[scale=0.33, viewport=82 25 534 388] {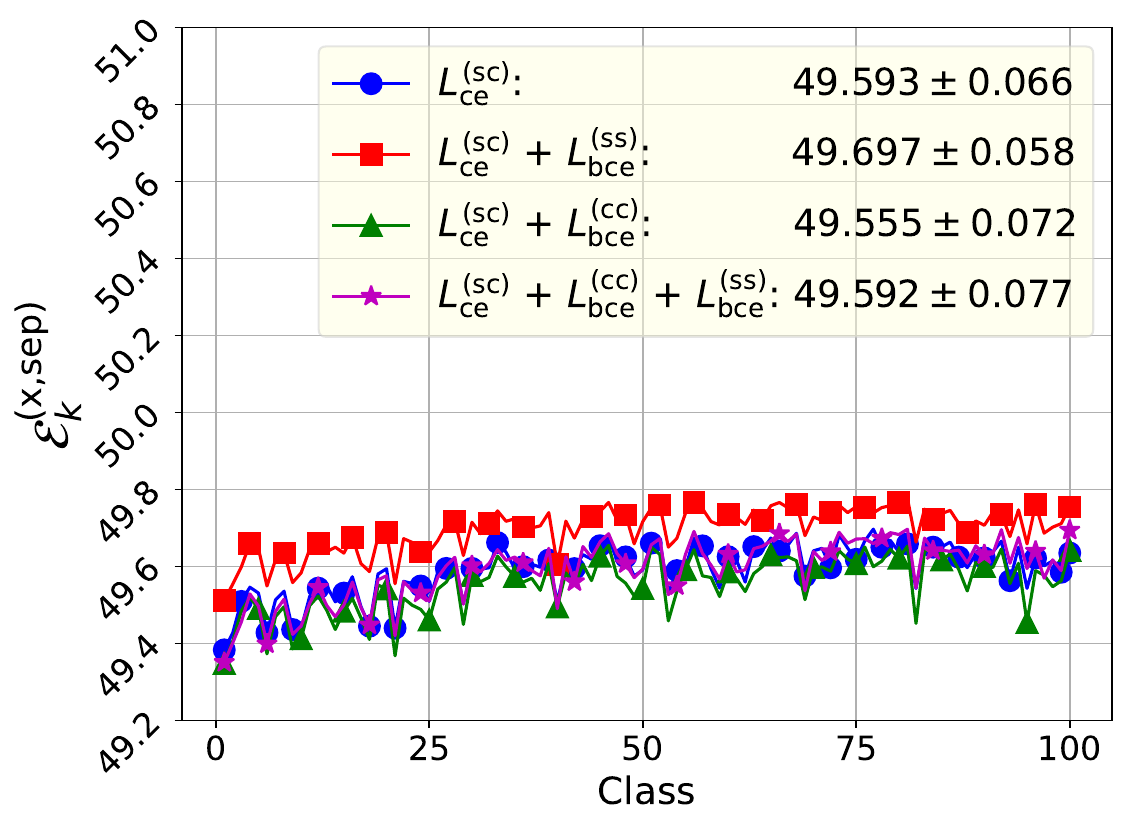}\hspace{-2pt}
    \includegraphics*[scale=0.33, viewport=82 25 535 388] {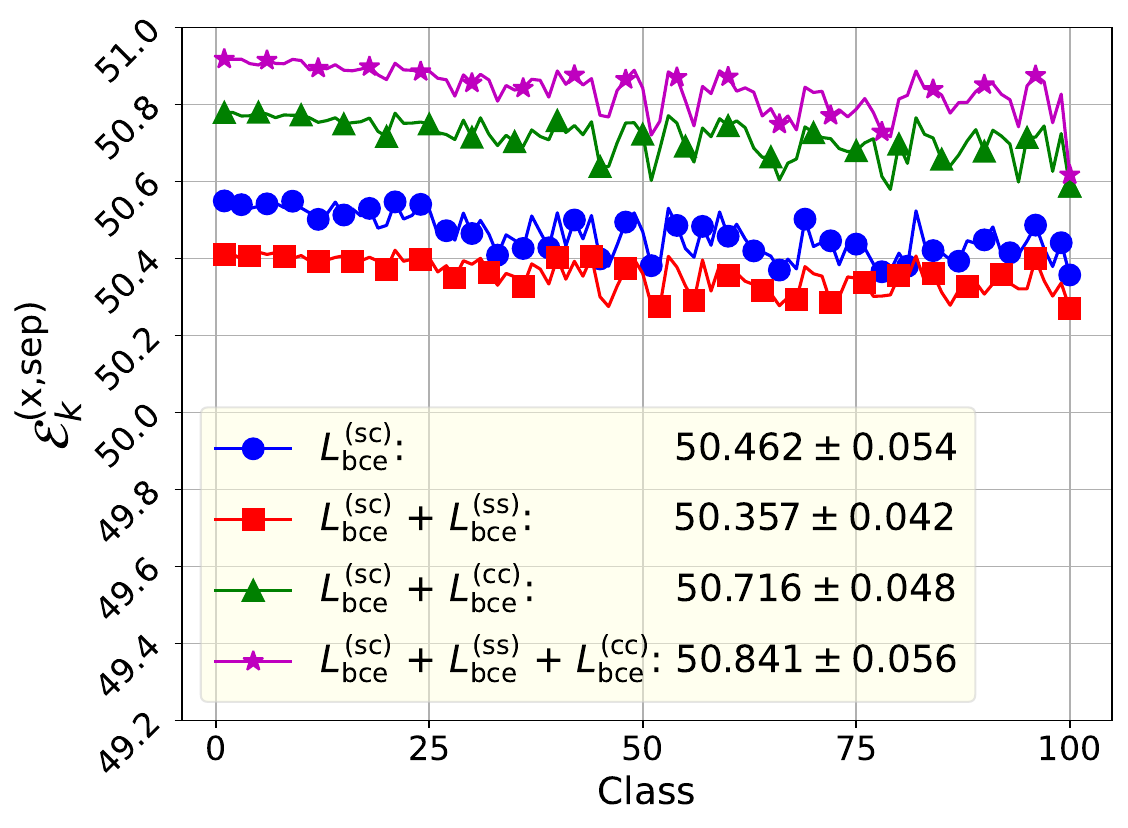}\\

    \includegraphics*[scale=0.33, viewport=8  25 534 388]{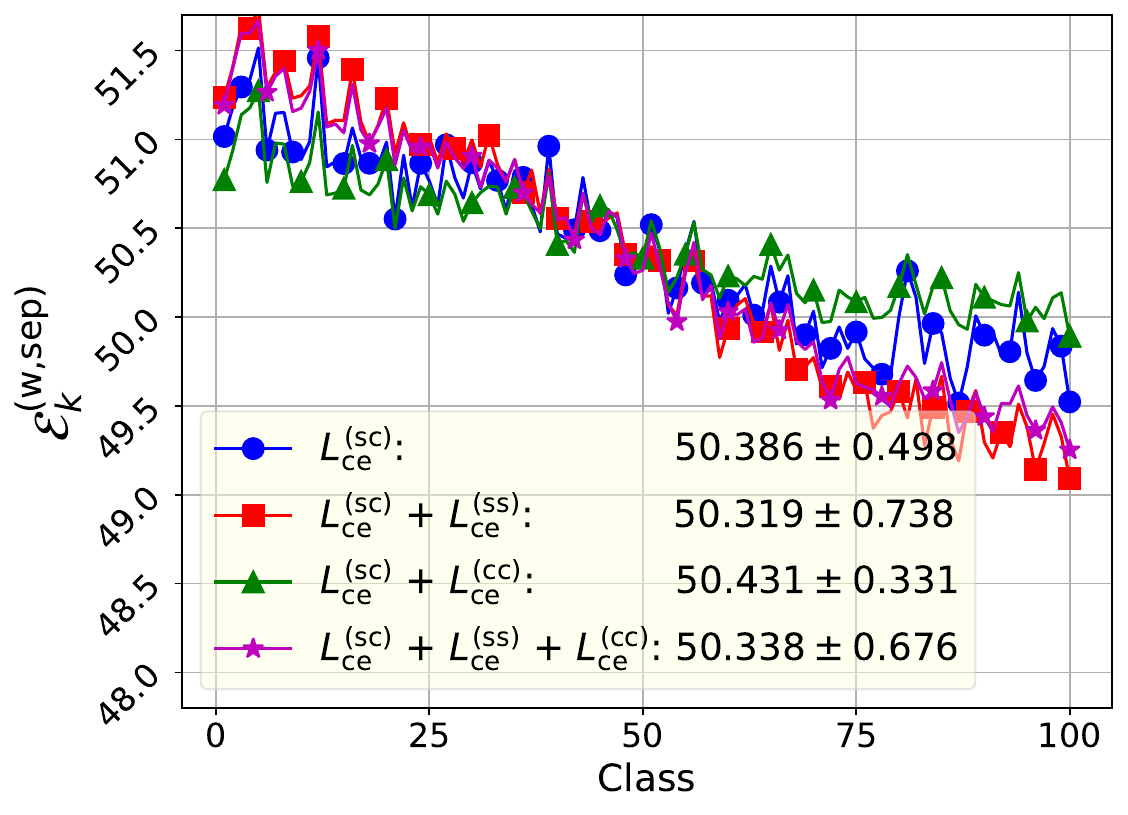}\hspace{-2pt}
    \includegraphics*[scale=0.33, viewport=82 25 535 388]{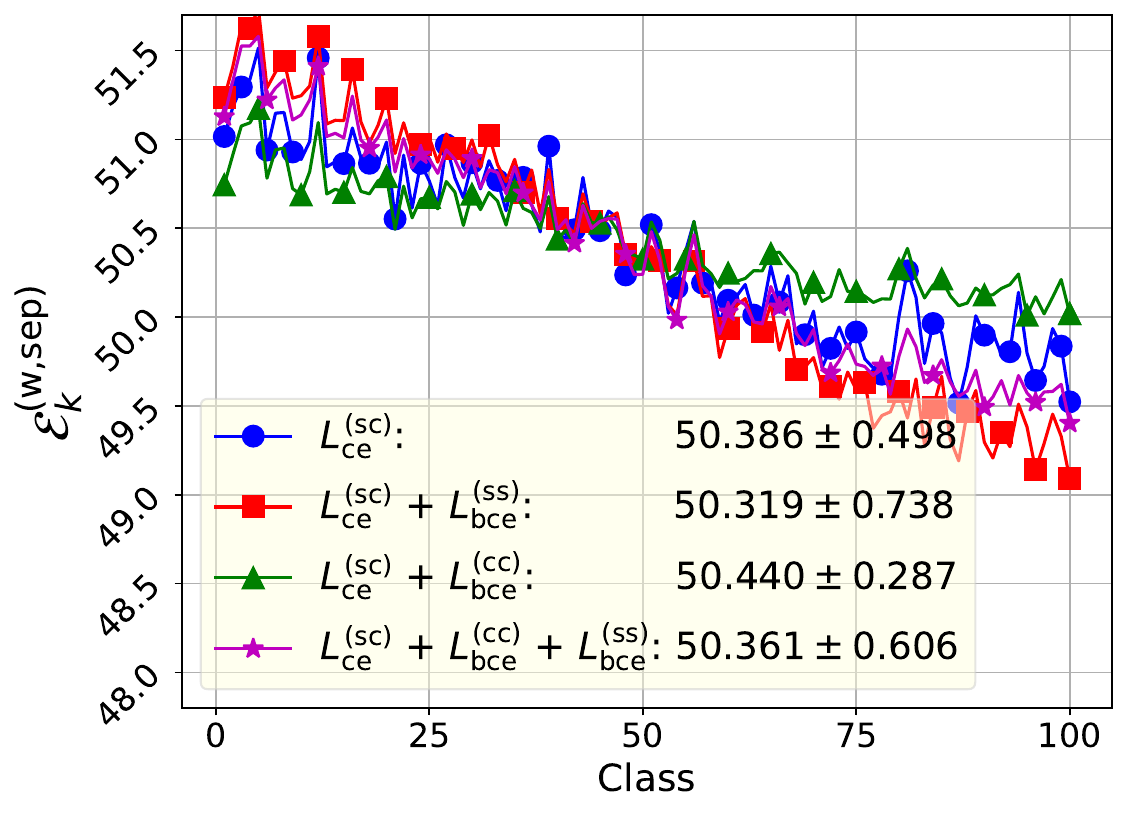}\hspace{-2pt}
    \includegraphics*[scale=0.33, viewport=82 25 535 388]{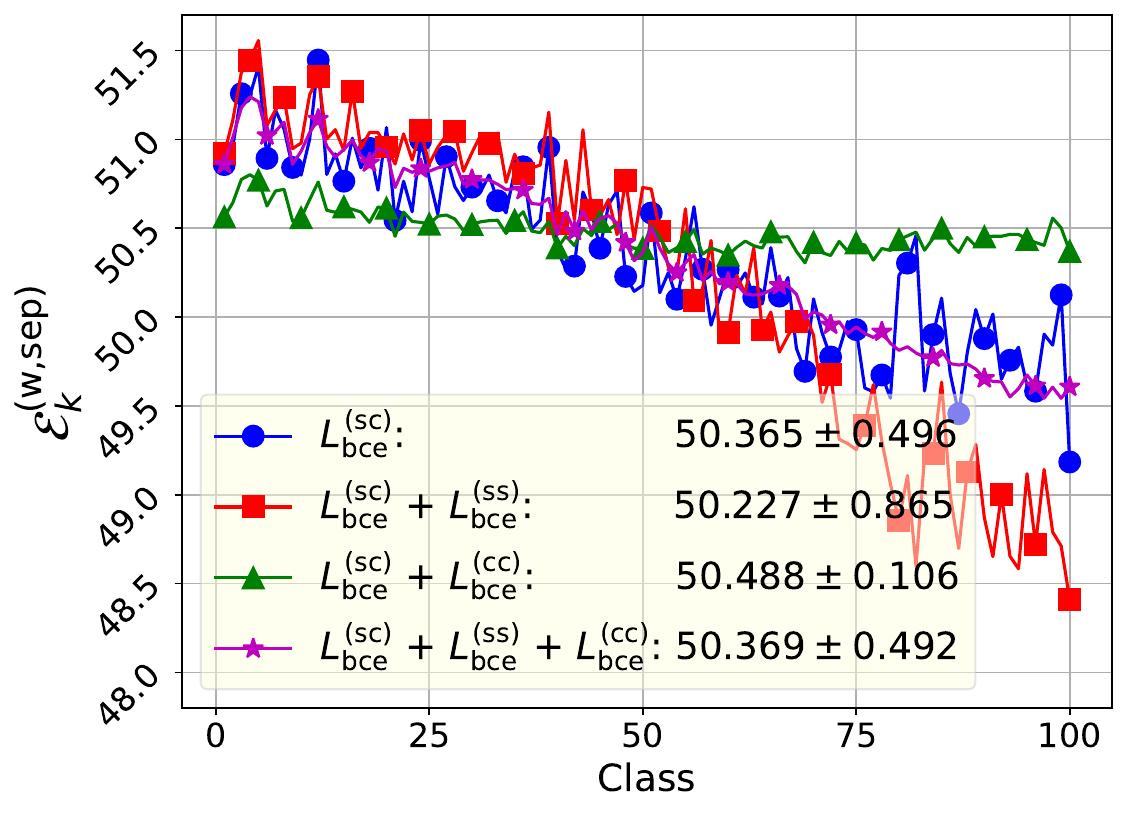}\\
    \caption{The intra-class compactness (top), inter-class separability (middle) of sample features, and separability (bottom) of classifier vectors on the training set of CIFAR100-LT (IF $=100$), with ResNet32 trained using the methods of CE-based learning (left), CE with BCE-based learning (middle), and BCE-based learning (right).}
    \label{supp_fig:CIFAR100_fu_comparsion_matrices_train}
\end{figure*}
\begin{figure*}[t]
    \centering
    \includegraphics*[scale=0.33, viewport=-2 25 534 388]{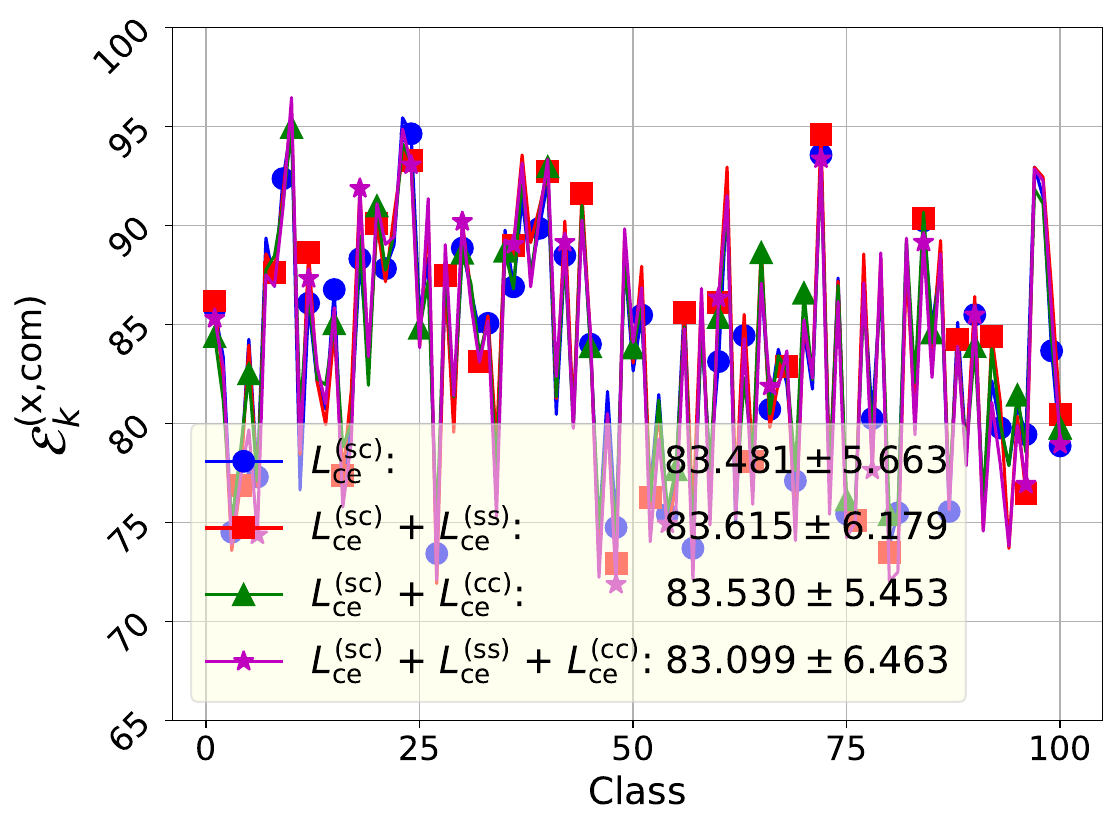}\hspace{-2.6pt}
    \includegraphics*[scale=0.33, viewport=80 25 535 388]{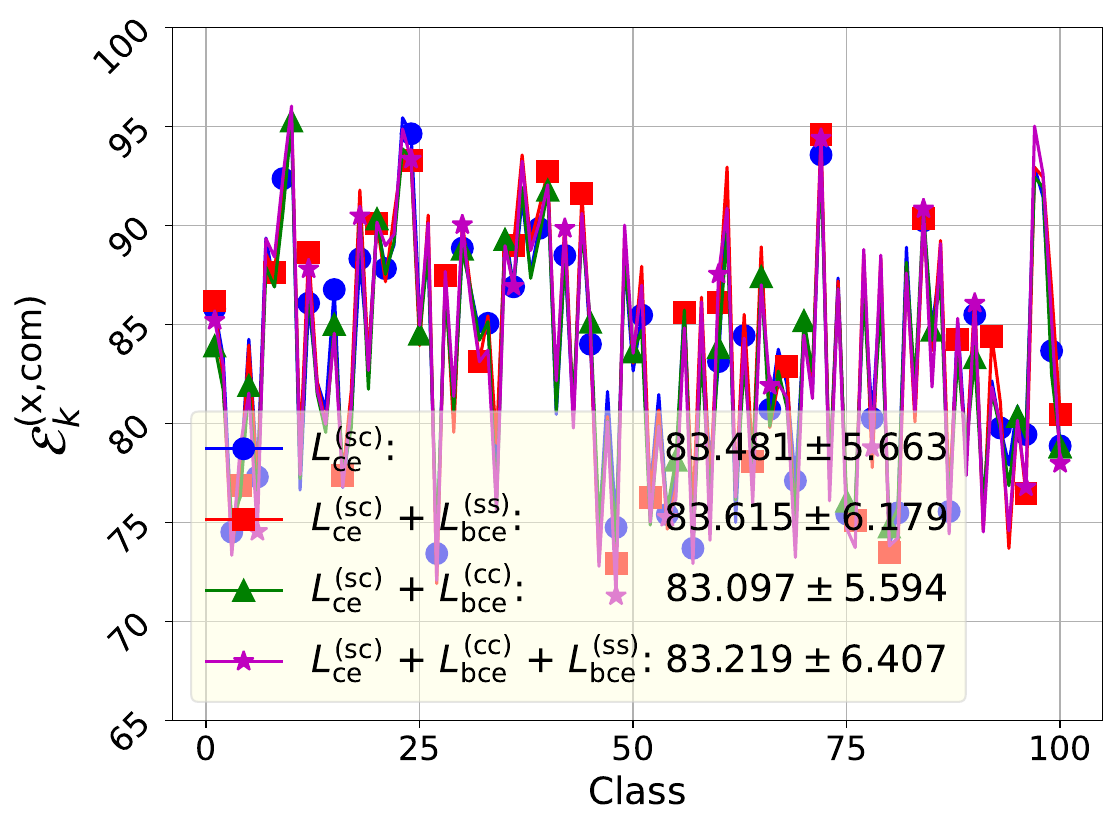}\hspace{-2pt}
    \includegraphics*[scale=0.33, viewport=80 25 535 388]{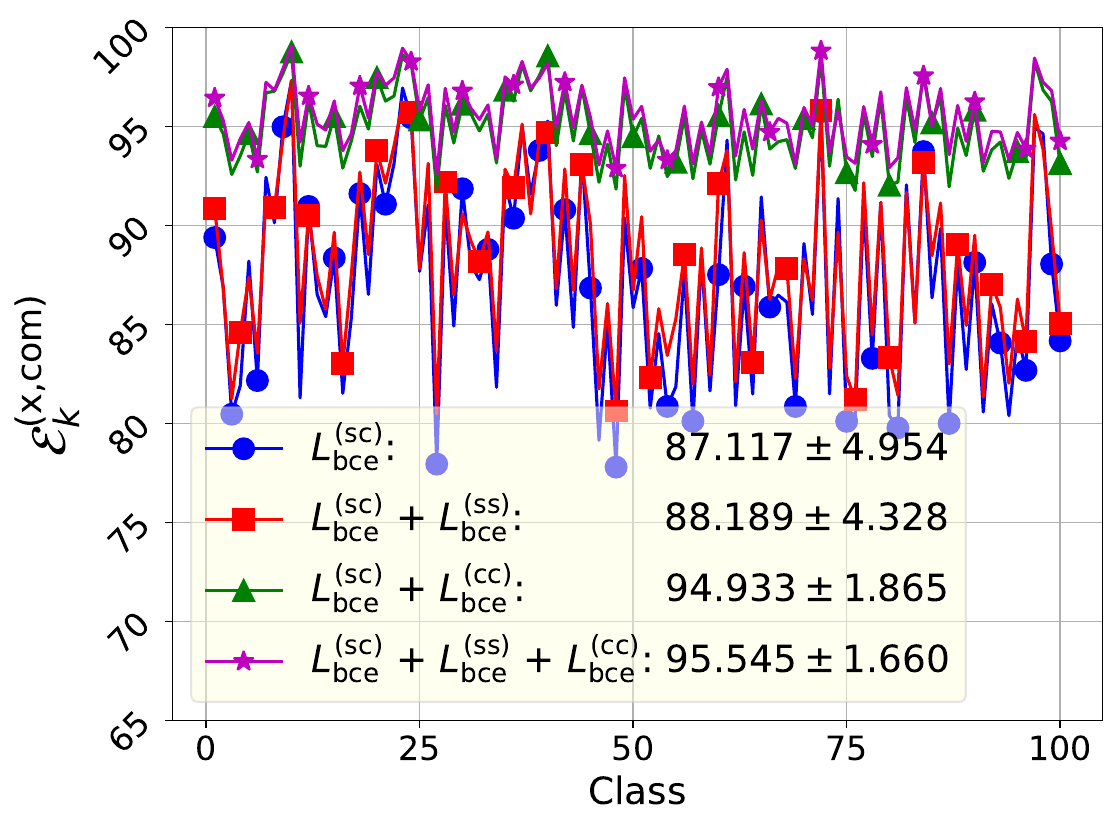}\\

    \includegraphics*[scale=0.33, viewport=8 8 534 388]  {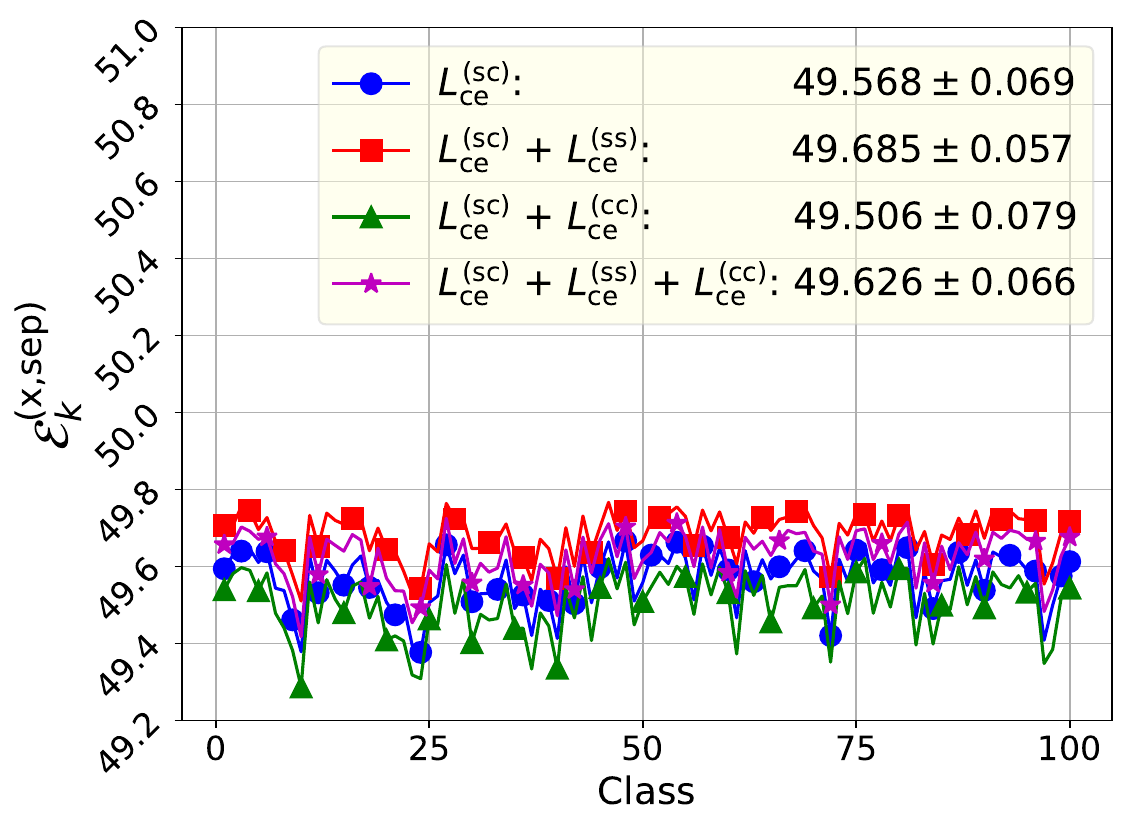}\hspace{-2pt}
    \includegraphics*[scale=0.33, viewport=82 8 534 388] {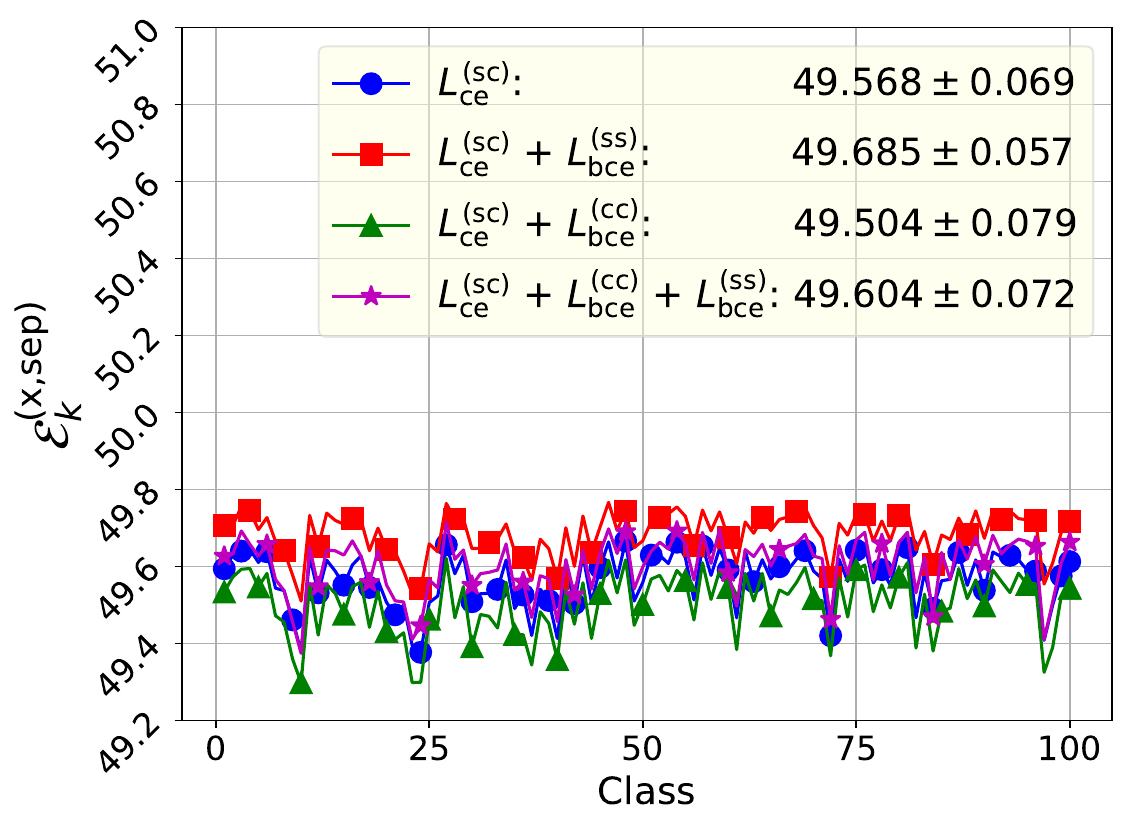}\hspace{-2pt}
    \includegraphics*[scale=0.33, viewport=82 8 535 388] {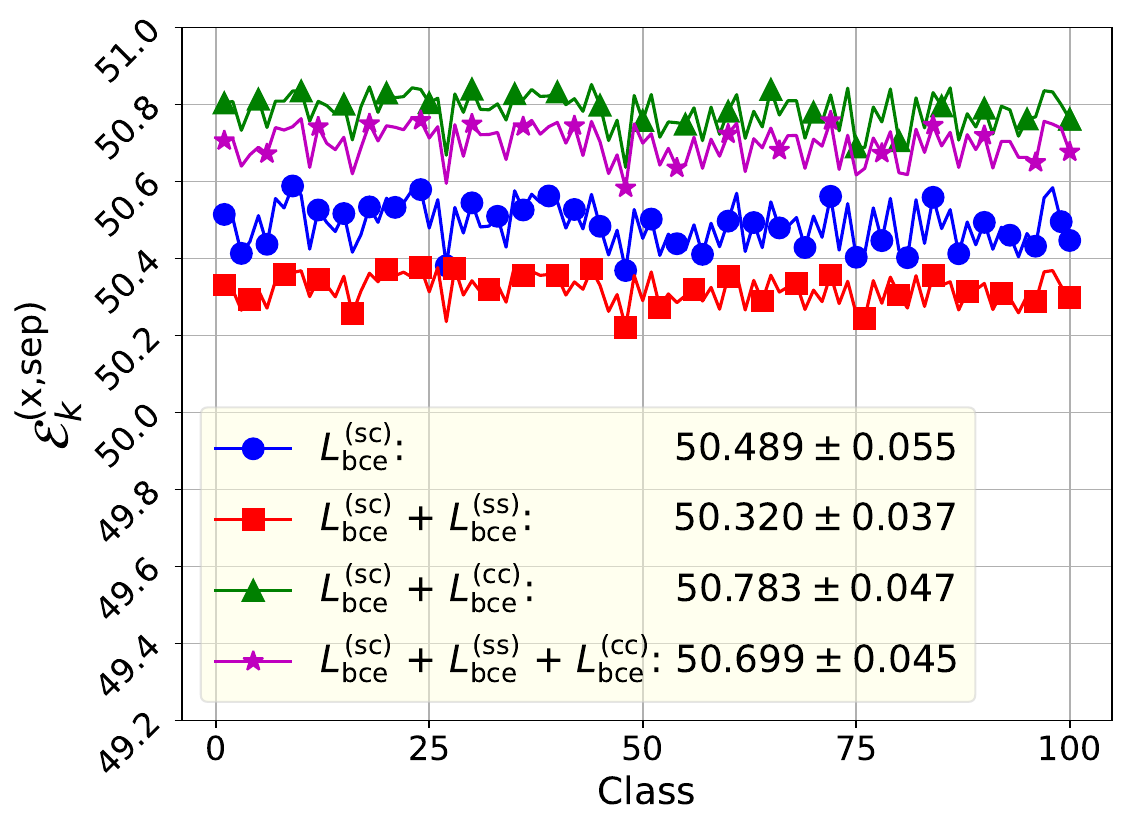}
    \caption{The intra-class compactness (top) and inter-class separability (bottom) of features on the test set of CIFAR100-LT (IF $=100$), with ResNet32 trained using the methods of CE-based learning (left), CE with BCE-based learning (middle), and BCE-based learning (right).}
    \label{supp_fig:CIFAR100_fu_comparsion_matrices_test}
\end{figure*}

\section*{Appendix G: More results for ablation experiments}\label{sec:supp_experiments_details}
\textbf{Parameter study.}\quad
BCE3S contains various hyper-parameters in its three BCE components, and we conduct the parameter study on CIFAR100-LT with IF $=100$.
For the re-sampling parameter $r$ in Eq.~(\ref{eq:sample_to_class_BCE}), we train ResNet32 using only the BCE joint learning $L_{\text{bce}}^{\text{(sc)}}$ with different $r$, while the results are shown in the top left of Fig \ref{fig:parameters_study}.
As $r$ decreases from $1.0$ to $0.4$, the accuracy on the \texttt{few} and \texttt{Medium} subsets increase from $15.5\%$ and $52.2\%$ to $17.4\%$ and $55.06\%$, respectively;
though the accuracy on the \texttt{Many} subset decreases, the total accuracy on the whole set has increased from $51.56\%$ to $52.88\%$.
When $r$ is below $0.4$, the accuracy drops significantly.
These results show that the hyper-parameter $r$ helps to balance the performances on the different classes in the LTR tasks.

For $\lambda_{\text{ss}}$ and temperature $\tau$ in the contrastive learning $L_{\text{bce}}^{\text{(ss)}}$,
we perform their parameter studies together using $L_{\text{bce}}^{\text{(sc)}} + \lambda_{\text{ss}}L_{\text{bce}}^{\text{(ss)}}$.
When increasing $\lambda_{\text{ss}}$ or decreasing $\tau$,
the contrastive learning is enhancing, which amplifies the imbalance effect of the long-tailed datasets and easily leads to training failure, as the bottom of Fig. \ref{fig:parameters_study} shows.

For the uniform learning $L_{\text{bce}}^{\text{(cc)}}$, a too small $\lambda_{\text{cc}}$ fails to enhance the classifier separability,
while a too large one would optimize the classifier too quickly, making it difficult to be further tuned with the updating features,
ultimately leading to a decreased LTR performance. As the top right of Fig. \ref{fig:parameters_study} shows, the best LTR corresponds to $\lambda_{\text{cc}}=1.25$ when using $L_{\text{bce}}^{\text{(sc)}} + \lambda_{\text{cc}}L_{\text{bce}}^{\text{(cc)}}$, significantly higher than that of CE ones.

\textbf{Separability and compactness.}\quad
In the ablation experiments, besides the CE and BCE-based tripartite synergistic learning, we also evaluate the combination of CE-based joint learning $L_{\text{ce}}^{\text{(sc)}}$ with BCE-based contrastive learning $L_{\text{bce}}^{\text{(ss)}}$ and uniform learning $L_{\text{bce}}^{\text{(cc)}}$.
The LTR accuracy of various learning methods on CIFAR100-LT with IF = 100 have been presented in Table \ref{table:ablation_study}, and we here further compare the classifier separability and features' compactness and separability for all the models.

Fig. \ref{supp_fig:CIFAR100_fu_comparsion_matrices_train} shows the separability and compactness of the different models on the training set.
We have compared the results of CE- and BCE-based tripartite synergistic learning in the experimental section of the main paper. 
We find and explain that both CE and BCE-based joint learning, $L_{\text{ce}}^{\text{(sc)}}$ and $L_{\text{bce}}^{\text{(sc)}}$, will learn imbalanced classifiers, but their learned features have significant differences. Because BCE's joint learning $L_{\text{bce}}^{\text{(sc)}}$ decouples the metric between any sample feature with $K$ classifier vectors $\{\bm w_k\}_{k=1}^K$, it avoids the secondary injection of classifier's imbalance effects into the feature learning. Therefore, compared to $L_{\text{ce}}^{\text{(sc)}}$, $L_{\text{bce}}^{\text{(sc)}}$ learns better features with higher compactness and separability.
Meanwhile, BCE-based contrastive learning $L_{\text{bce}}^{\text{(ss)}}$ can enhance the compactness of features, and BCE-based uniform learning $L_{\text{bce}}^{\text{(cc)}}$ effectively balances the separability of classifiers.
However, the two CE-based learning modes, $L_{\text{ce}}^{\text{(ss)}}$ and $L_{\text{ce}}^{\text{(cc)}}$, have no significant effect in this regard.

Here, we further found that although BCE-based contrastive learning $L_{\text{bce}}^{\text{(ss)}}$ and uniform learning $L_{\text{bce}}^{\text{(cc)}}$ can improve the properties of features and classifiers, they do not show these advantages when combined with CE-based joint learning $L_{\text{ce}}^{\text{(sc)}}$.
As the left and middle columns in Fig. \ref{supp_fig:CIFAR100_fu_comparsion_matrices_train} show, when $L_{\text{bce}}^{\text{(ss)}}$ and $L_{\text{bce}}^{\text{(cc)}}$ are combined with $L_{\text{ce}}^{\text{(sc)}}$, the final classifier separability and features' compactness and separability are almost identical to that obtained by combining $L_{\text{ce}}^{\text{(ss)}}$, $L_{\text{ce}}^{\text{(cc)}}$ with $L_{\text{ce}}^{\text{(sc)}}$.
These results suggest that CE-based joint learning has inherent limitations in LTR tasks, which is difficult to overcome completely by external re-balancing techniques.
We believe that this defect of CE comes from its Softmax function, which couples $K$ metrics related to the imbalanced classifiers on its denominator.

Fig. \ref{supp_fig:CIFAR100_fu_comparsion_matrices_test} shows the features' compactness and diversity of models trained with different learning methods on the test set of CIFAR100-LT with IF = 100. We draw similar conclusions from these results as we did on the training set: BCE-based joint learning $L_{\text{bce}}^{\text{(sc)}}$ can learn better sample features than CE-based one $L_{\text{ce}}^{\text{(sc)}}$, and its contrastive learning $L_{\text{bce}}^{\text{(ss)}}$ and uniform learning $L_{\text{bce}}^{\text{(cc)}}$ can further effectively improve the feature compactness, but they cannot improve the feature properties of CE-based joint learning $L_{\text{ce}}^{\text{(sc)}}$. These results also reflect that BCE3S has well generalization.

\begin{figure*}[t]
    \centering
    \includegraphics[width=0.96\linewidth]{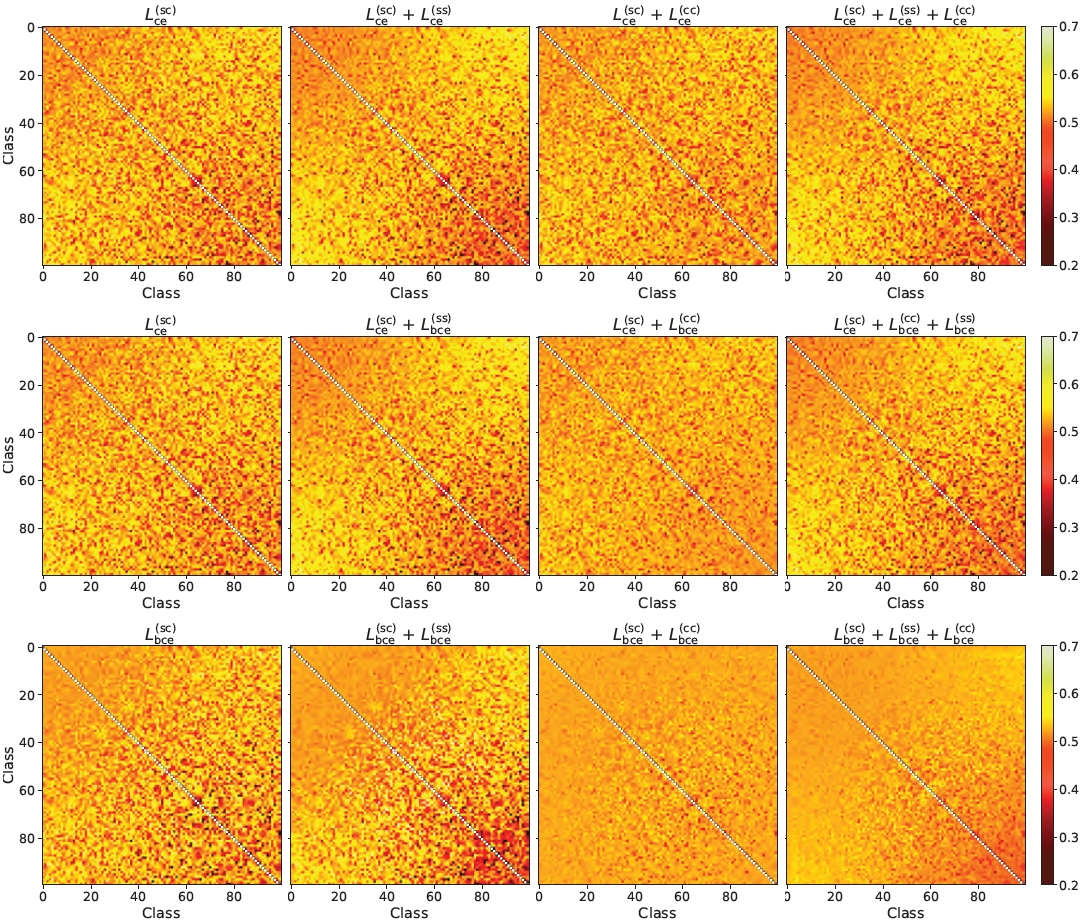}
    \caption{
        The visualizations of similarity matrices of classifier trained by different learning configurations: CE-based learning (top), CE with BCE-based learning (middle), and BCE-based learning (bottom).
    }
    \label{supp_fig:CIFAR100_separability_heatmap}
\end{figure*}

\textbf{Separability matrix of classifiers.}\quad
We also compute the separability matrix $S = \big[s_{jk}\big]$ for the classifier, where
\begin{align}
s_{jk} =
\left\{
\begin{array}{cc}
    1& j = k,\\
    \frac{1-\cos(\hat{\bm w}_j, \hat{\bm w}_k)}{2}& j\neq k.
\end{array}
\right.
\end{align}
Fig. \ref{supp_fig:CIFAR100_separability_heatmap} shows the separability matrix of classifiers in models trained by different CE and BCE-based methods on on CIFAR100-LT with IF = 100. In each matrix diagram, the upper left area reflects the classifier separability among head classes, the lower left and upper right areas reflect the classifier separability between head and tail classes, and the lower right area reflects the classifier separability among tail classes.
Except for the elements on the diagonal, the brighter points indicate the higher separability between the two classifier vectors, while the darker points indicate the lower separability. From the first column of the figure, one can find that in each subfigure, the lower left and upper right areas are the brightest, followed by the upper left area, while the lower right area has more obvious black points, indicating that the BCE and CE-based joint learning, $L_{\text{ce}}^{\text{(sc)}}$ and $L_{\text{bce}}^{\text{(sc)}}$, achieve the highest classifier separability between the head and tail classes, followed by the classifier separability among the head classes, while the classifier separability among the tail classes is the worst. 

From the second column in Fig. \ref{supp_fig:CIFAR100_separability_heatmap}, one can find that when the two kind of contrastive learning are respectively added, the imbalance of the classifier separability of the models are further exacerbated, with the lower left and upper right area of each subfigure appearing brighter and the lower right area appearing darker.
The third column shows the separability matrix of the classifier trained by the combination of joint learning and uniform learning. One can find that the combination of $L_{\text{bce}}^{\text{(sc)}}$ and $L_{\text{bce}}^{\text{(cc)}}$ effectively improves the separability of the classifier, and the color of the entire separability matrix diagram is more balanced among its different parts. However, combined with CE-based joint learning $L_{\text{ce}}^{\text{(sc)}}$, neither BCE-based uniform learning $L_{\text{bce}}^{\text{(cc)}}$ nor CE-based one $L_{\text{ce}}^{\text{(cc)}}$ effectively balances the classifiers, and the colors of the separability matrix on the two diagonal directions still exhibit significant differences. In the fourth column, when contrastive learning is added to the model training again, the color of the separability matrix corresponding to CE and BCE in different parts becomes more uneven or re-uneven, indicating an imbalanced classifier.

\begin{figure*}[t]
    \centering
    \small
    \includegraphics[width=1.0\linewidth]{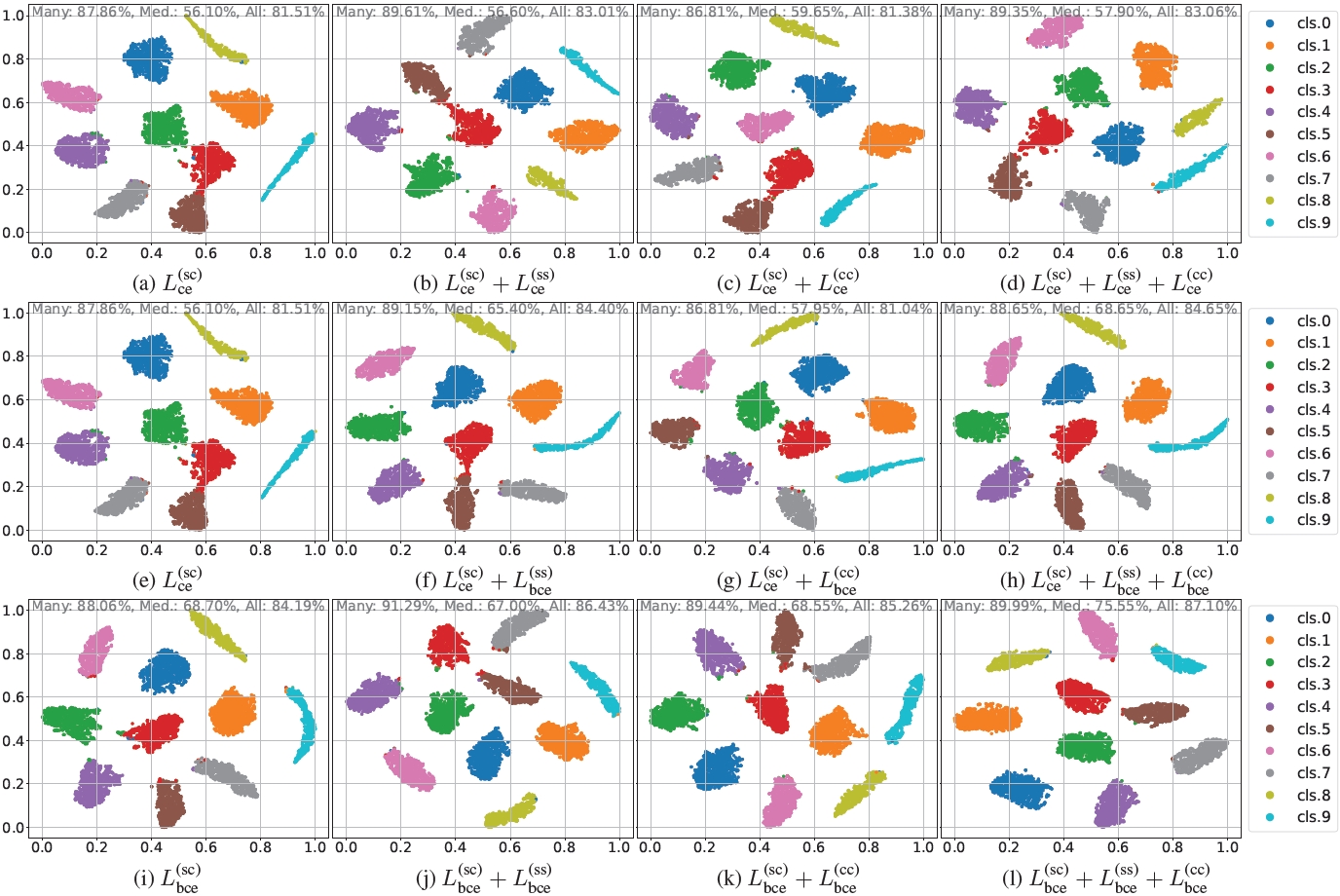}
    \caption{
        Feature distribution on the test set of CIFAR10-LT (IF = 100) with different learning configurations: CE-based learning (top), CE with BCE-based learning (middle), and BCE-based learning (bottom).
    }
    \label{supp_fig:CIFAR10_fu_comparsion_feature_three_col}
\end{figure*}

\textbf{Feature distribution in t-SNE.}\quad
To further compare the features of CE and BCE-based learning modes, we trained models on CIFAR10-LT with IF of 100, then, using t-SNE, we visually show their feature distributions of all 10 classes on the test set, as Fig.~\ref{supp_fig:CIFAR10_fu_comparsion_feature_three_col} shows.
In the figure, the first row shows the features of model trained by various learning combination of three CE-based learning modes, i.e., $L_{\text{ce}}^{\text{(sc)}}$, $L_{\text{ce}}^{\text{(ss)}}$, and $L_{\text{ce}}^{\text{(cc)}}$; the second row shows that of CE-based joint learning with BCE-based contrastive learning and uniform learning, i.e., $L_{\text{ce}}^{\text{(sc)}}$ with $L_{\text{bce}}^{\text{(ss)}}$, $L_{\text{bce}}^{\text{(cc)}}$; the third row shows that of BCE-based three learning modes, i.e., $L_{\text{bce}}^{\text{(sc)}}$, $L_{\text{bce}}^{\text{(ss)}}$, and $L_{\text{bce}}^{\text{(cc)}}$.
Similarly, the sample features learned by BCE-based joint learning $L_{\text{bce}}^{\text{(sc)}}$ exhibit better compactness and separability than that learned by CE one $L_{\text{ce}}^{\text{(sc)}}$, and $L_{\text{bce}}^{\text{(ss)}}$ and $L_{\text{bce}}^{\text{(cc)}}$ can further effectively improve the properties of features.
In contrast, the compactness and separability of sample features learned by CE-based joint learning $L_{\text{ce}}^{\text{(sc)}}$ are poor, and their corresponding contrastive learning $L_{\text{ce}}^{\text{(ss)}}$ and uniform learning $L_{\text{ce}}^{\text{(cc)}}$ cannot effectively improve the feature properties, which were only slightly improved by BCE-based contrastive learning $L_{\text{bce}}^{\text{(ss)}}$ and uniform learning $L_{\text{bce}}^{\text{(cc)}}$.

\section*{Appendix H: More results about Comparison BCE with SOTA}\label{sec:supp_comparsion_with_SOTA}
On \textbf{ImageNet-LT}, we apply BCE3S to train ResNet50 and ResNeXt50 by adopting a similar strategy used in DSCL~\cite{xuan2024_aaai_dscl}. We compare BCE3S with various state-of-the-art (SOTA) methods, including CB-Focal~\cite{cui2019class}, RISDA~\cite{chen2022imagine_RISDA}, LDAM~\cite{cao2019learning_margin_loss}, KCL~\cite{kang2021exploring}, TSC~\cite{li_2022_targeted_TSC}, GCL~\cite{li2022logits_adjust_ce}, GCL+H2T~\cite{li_2024_feature_fusion}, BCL~\cite{zhu_2022_balanced_BCL}, DiffuLT~\cite{DiffuLT2024}, BCL+DODA~\cite{wang_iclr2024_kill}, DiffuLT+RIDE~\cite{DiffuLT2024}, DSCL~\cite{xuan2024_aaai_dscl}, ProCo~\cite{du_2024_probabilistic_proco}, ETF+DR~\cite{yang_2022inducing_nc}+DR~\cite{Kang_2020_Decoupling}, FCL~\cite{kang2021exploring}, RBL~\cite{Peifeng_2023_nc_LTR}, NC-DRW~\cite{liu_2023_inducing}, NC-cRT~\cite{liu_2023_inducing}, Meta Softmax~\cite{ren2020balanced_ms}, PaCo~\cite{cui_2021_parametric_paco}, GLMC~\cite{du2023glmc}, SSD~\cite{li_2021_self_SSD}, GLMC+MN~\cite{2022_maxnorm}, and GLMC+MS~\cite{ren2020balanced_ms}.
We have presented the LTR results of ResNet50 in the main body of the paper, and we here present the results of ResNeXt50.

Table~\ref{table:supp_benchmark_on_imagenet_lt_rx50} provides a detailed comparison of these methods using ResNeXt50 on ImageNet-LT. Although BCE3S achieved only one optimal result on the three test subsets (\texttt{Many}, \texttt{Medium}, and \texttt{Few}), it outperformed all competing methods in terms of overall accuracy on the whole test set. Specifically, BCE3S achieved $58.54\%$ with ResNeXt50, surpassing prior SOTA results like GLMC~\cite{du2023glmc}, and ProCo~\cite{du_2024_probabilistic_proco}.

\begin{figure}[t]
    \centering

    \includegraphics*[scale=0.4]{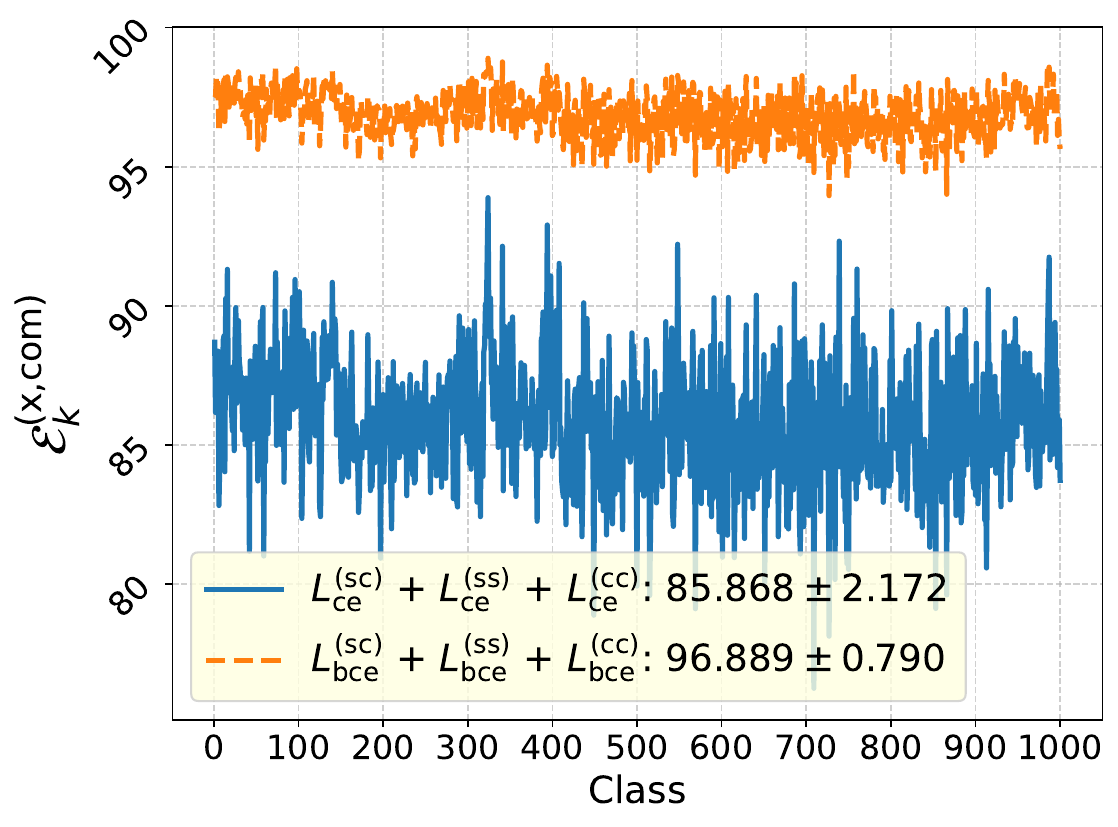} \\
    \includegraphics*[scale=0.4]{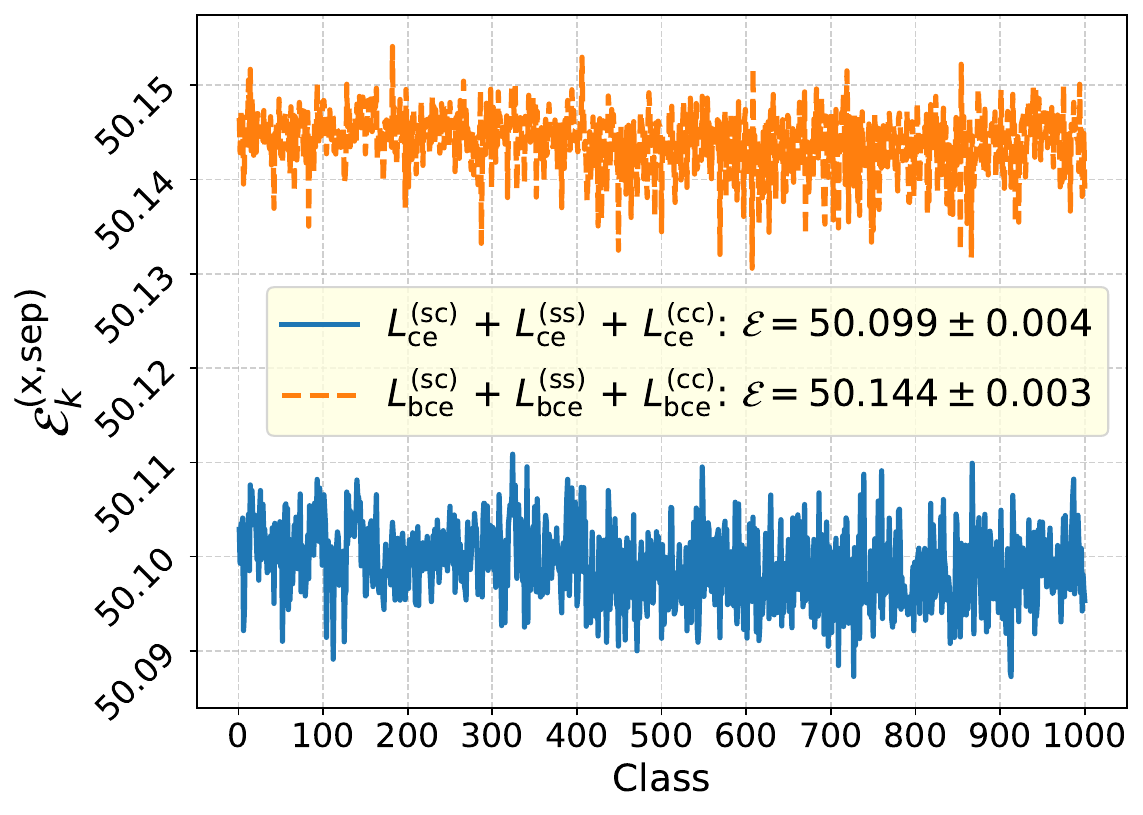} \\ 
    \includegraphics*[scale=0.4]{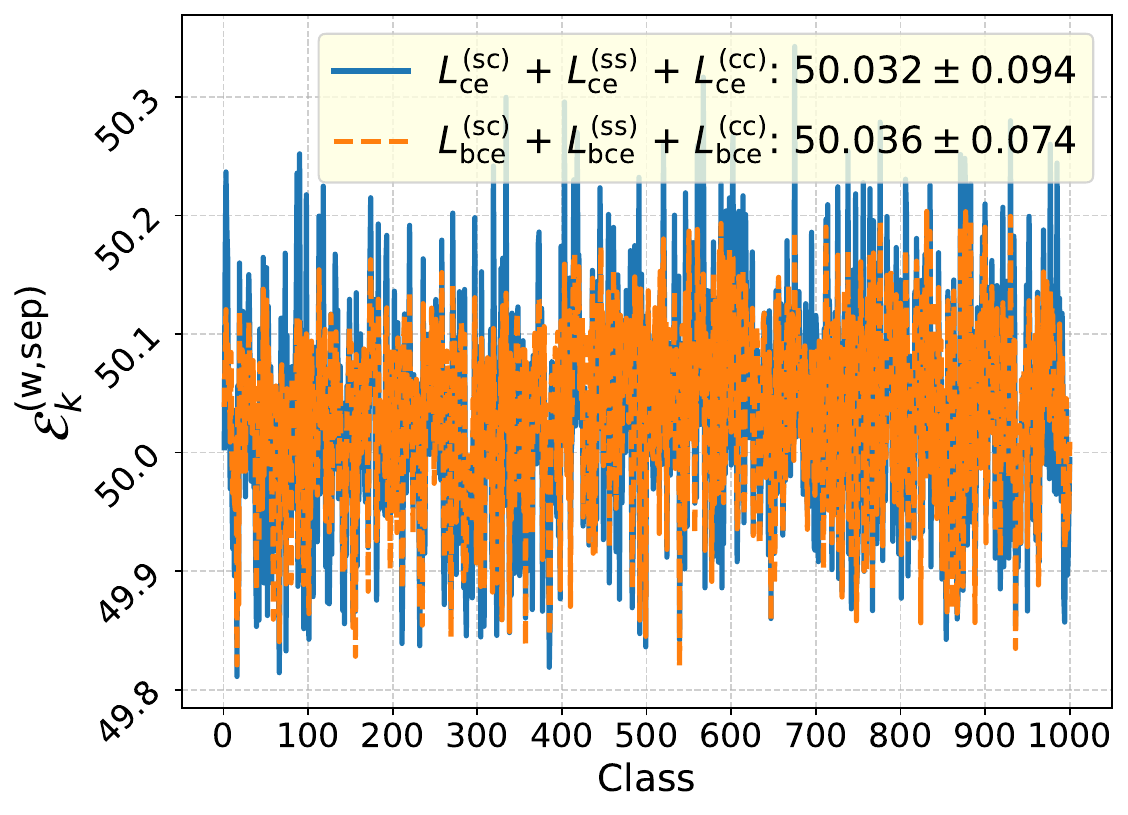} \\
    \caption{The intra-class compactness (left), inter-class separability (middle) of features and inter-class separability of classifier on the train set of ImageNet-LT with ResNet50.}
    \label{supp_fig:imagenet_lt_comparsion_matrices_train}
\end{figure}

\begin{table}[!t]
    \centering
    \small
    \arrayrulecolor{black}
    \ADLnullwidehline
    \setlength{\tabcolsep}{1.2mm}{
        \begin{tabular}{l|c|cccc}
        \arrayrulecolor{black}\hline
        \multirow{2}{*}{Methods} & \multirow{2}{*}{$\mathcal M$}    & \multicolumn{4}{c}{ImageNet-LT}          \\
                                &                              & \texttt{Many}       & \texttt{Med.}       & \texttt{Few}   & All    \\
        \cline{1-4}\arrayrulecolor{black}\cline{5-6}
        ETF+DR\tiny{NeurIPS'22} & \multirow{17}{*}{RX50} & -          & -          & -     & 44.70  \\
        FCL\tiny{ICLR'21}       &                        & 61.40      & 47.00      & 28.20 & 49.80  \\
        KCL\tiny{ICLR'21}       &                        & 62.40      & 49.00      & 29.50 & 51.50  \\
        TSC\tiny{CVPR'22}       &                        & 63.50      & 49.70      & 30.40 & 52.40  \\
        RBL\tiny{ICML'23}       &                        & 64.80      & 49.60      & 34.20 & 53.30  \\
        NC-DRW\tiny{AISTATS'23} &                        & 67.10      & 49.70      & 29.00 & 53.60  \\
        NC-cRT\tiny{AISTATS'23} &                        & 65.60      & 51.20      & 35.40 & 54.20  \\
        Meta Soft.\tiny{NeurIPS'20}  &                        & 65.80      & 53.20      & 34.10 & 55.40  \\
        PaCo\tiny{ICCV'21}      &                        & 64.40      & \underline{55.70}      & 33.70 & 56.00  \\
        GLMC\tiny{CVPR'23}      &                        & 70.10      & 52.40      & 30.40 & 56.30  \\
        SSD\tiny{ICCV'21}       &                        & 66.80      & 53.10      & 35.40 & 56.60  \\
        GLMC+MN\tiny{CVPR'22}   &                        & 60.80      & \textbf{55.90} & \textbf{45.50} & 56.70  \\
        BCL\tiny{CVPR'22}       &                        & 67.90      & 54.20      & 36.60 & 57.10  \\
        GLMC + MS \tiny{NeurIPS'20}     &                        & 64.76      & 55.67      & \underline{42.19} & 57.21  \\
        ProCo\tiny{TPAMI'24}    &                        & -          & -          & -     & \underline{58.00}  \\
        BCE3S                   &                        & \textbf{71.13} & 54.98      & 36.78 & \textbf{58.54}  \\
        \arrayrulecolor{black}\cline{1-4}\arrayrulecolor{black}\cline{5-6}
        \end{tabular}
    }
    \arrayrulecolor{black}
    \caption{Benchmark results of top-1 accuracy (\%) on ImageNet-LT. ResNeXt50 (RX50), trained by BCE3S, achieves the best results on the test set.}
    \label{table:supp_benchmark_on_imagenet_lt_rx50}
    \vspace{-8pt}
\end{table}

\section*{Appendix I: Further analysis on ImageNet-LT}
\textbf{Compactness and separability. }\quad
Fig.~\ref{supp_fig:imagenet_lt_comparsion_matrices_train} presents three metrics for ResNet50 trained on ImageNet-LT across all 1000 classes. For intra-class feature compactness, BCE3S demonstrates superior performance with an average of 96.889 and standard deviation of 0.790, compared to CE3S which achieves 85.868 with 2.172 standard deviation. This indicates BCE3S produces significantly more uniform compactness across both head and tail classes. Similarly, for inter-class feature separability, BCE3S exhibits enhanced uniformity across the class distribution, with 50.144 compared CE's 50.099. Regarding classifier separability, despite the increased complexity from the large number of classes, BCE3S still outperforms CE3S consistently. These results provide compelling evidence that BCE effectively addresses the intrinsic limitations of CE in long-tailed recognition tasks, particularly in maintaining balanced feature representations across the entire class distribution.